%% file: main.tex
\documentclass[conference]{IEEEtran}
\IEEEoverridecommandlockouts

\usepackage{cite}
\usepackage{amsmath,amssymb,amsfonts}
\usepackage{algorithmic}
\usepackage{graphicx}
\usepackage{textcomp}
\usepackage{xcolor}
\input{include/includes}
\input{include/setup}

\usepackage{kamomargins}
\kamoInit{}
\kamoSetMargins{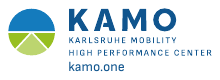}
\kamoSetIEEEFoot{2025}
\kamoSetIEEEHead{Dudzik, L., Roschani, M., Sielemann, A., Trampert, K., Ziehn, J., Beyerer, J., \& Neumann, C. (2025, September). Physically-Based Simulation of Automotive LiDAR. In 2025 IEEE International Automated Vehicle Validation Conference (IAVVC)}
{10.1109/IAVVC61942.2025.11219516}

\begin{document}

\title{Physically-Based Simulation of Automotive LiDAR
\thanks{This publication was written within the framework of the KAMO: Karlsruhe Mobility High Performance Center (www.kamo.one), in the context of the AVEAS research project (www.aveas.org), funded by the German Federal Ministry for Economic Affairs and Climate Action (BMWK) within the program ``New Vehicle and System Technologies'', and supported by the Fraunhofer Internal Programs under Grant No. 40-00736 within the ``ALBACOPTER\registered'' project (www.albacopter.fraunhofer.de).}
}

\author{%
\IEEEauthorblockN{
L. Dudzik\IEEEauthorrefmark{2},
M. Roschani\IEEEauthorrefmark{1}, A. Sielemann\IEEEauthorrefmark{1},
K. Trampert\IEEEauthorrefmark{2},
J. Ziehn\IEEEauthorrefmark{1},
J. Beyerer\IEEEauthorrefmark{1}\IEEEauthorrefmark{3} and C. Neumann\IEEEauthorrefmark{2}}
\IEEEauthorblockA{\IEEEauthorrefmark{2}Light Technology Institute (LTI),
Karlsruhe Institute of Technology (KIT), 76131 Karlsruhe, Germany\\
Email: \{leonhard.dudzik, klaus.trampert\}@kit.edu}
\IEEEauthorblockA{\IEEEauthorrefmark{3}Vision and Fusion Laboratory (IES),
Karlsruhe Institute of Technology (KIT), 76131 Karlsruhe, Germany}
\IEEEauthorblockA{\IEEEauthorrefmark{1}Fraunhofer IOSB, 
Email: \{anne.sielemann, masoud.roschani, jens.ziehn\}@iosb.fraunhofer.de}%
}

\maketitle

\begin{abstract}
We present an analytic model for simulating automotive time-of-flight (ToF) LiDAR that includes blooming, echo pulse width, and ambient light, along with steps to determine model parameters systematically through optical laboratory measurements.
The model uses physically based rendering (PBR) in the near-infrared domain. It assumes single-bounce reflections and retroreflections over rasterized rendered images from shading or ray tracing, including light emitted from the sensor as well as stray light from other, non-correlated sources such as sunlight. Beams from the sensor and sensitivity of the receiving diodes are modeled with flexible beam steering patterns and with non-vanishing diameter.
Different (all non-real time) computational approaches can be chosen based on system properties, computing capabilities, and desired output properties.

Model parameters include system-specific properties, namely the physical spread of the LiDAR beam, combined with the sensitivity of the receiving diode; the intensity of the emitted light; the conversion between the intensity of reflected light and the echo pulse width; and scenario parameters such as environment lighting, positioning, and surface properties of the target(s) in the relevant infrared domain. System-specific properties of the model are determined from laboratory measurements of the photometric luminance on different target surfaces aligned with a goniometer at 0.01° resolution, which marks the best available resolution for measuring the beam pattern. 
The approach is calibrated for and tested on two automotive LiDAR systems, the Valeo Scala Gen. 2 and the Blickfeld Cube 1. Both systems differ notably in their properties and available interfaces, but the relevant model parameters could be extracted successfully.

\end{abstract}


\section{Motivation and State of the Art}

\IEEEPARstart{T}{he} provision of realistic data for training and testing of safety-critical systems has received additional attention lately. Progress in highly automated or unmanned transportation systems on and off roads and in the air has enabled use cases that demand high safety levels in complex operational design domains \cite{eisemann2024joint}. This progress is fueled, to a large extent, by advanced data-driven models based on deep learning. For these, data currently serve both as the primary means of defining functions and as a crucial means of evaluating and validating these functions. Accordingly, new legislations and standards \cite{genovesi2025evaluating}, in particular the European AI Act \cite{aia}, set high standards for the quality of such data. As real-world data acquisition and annotation is a laborious, time-consuming, costly, and error-prone task, synthetic data (mainly from simulations, generative AI, or a combination thereof) are pursued as a plausible alternative. However, to ensure that such data can adequately replace real data with the confidence required for the testing of safety-critical systems, it must be demonstrated that the domain gap between these data and the real world is sufficiently small and sufficiently well understood. Progress has been made in particular for camera-based imaging \cite{SYNTHIAdataset,gta5SemanticSegmentation,gta5objectDetection,vKITTI,vKITTI2,bullinger2020photogrammetry,ziehn2020imaging,sielemann2024synset,synset_signset_ger_sielemann_2024}, which has considerable similarities to LiDAR optics, besides important differences.

The synthetic generation of LiDAR data has also received attention, albeit to a still lesser degree. One of the earliest detailed descriptions using rasterization rendering as a prior for LiDAR simulation is given in \cite{peinecke2008lidar}, where real-time shader artifacts from OpenGL were utilized to generate dense point clouds for airborne sensor data. Applications for using synthetic, shader-based LiDAR data for unmanned aerial vehicle applications are given in \cite{riordan2021lidar}. Examples of detailed and application-specific models include \cite{yang2022comprehensive} for a detailed per-ray model for vegetation monitoring, \cite{browning20123d} for ground-based volumetric vegetation perception, or \cite{abdallah2012wa} for water appearance. An approach to augment real point clouds of static road environments with synthetic traffic objects is given in \cite{fang2020augmented}. \cite{manivasagam2020lidarsim} provides a data-driven model based on raycasting and the subsequent application of a U-Net ML model to bridge the domain gap by introducing realistic raydrops (i.e., rays not yielding echoes) based on range, incident angle, intensities, and additional features. This work is expanded in \cite{li2023pcgen} with a modified model for raydrop estimation and surface approximation, and in \cite{manivasagam2023towards} where an exhaustive description of various effects is provided, including (beyond raydrops) multiple echoes, spurious returns from indirect paths and blooming, noise, and time delays between points, providing a differentiated analysis of the impacts of each aspect on the domain gap. Even more data-driven approaches lie in the generative AI-based creation of point clouds, where \cite{zyrianov2024lidardm} gives an overview of approaches and presents a novel diffusion-based model to produce temporally consistent point clouds for traffic scenes.

For this paper, we focus on two aspects that received lesser attention in previous work: The metrological integration of optical phenomena, especially blooming (\cref{sec:intro-blooming}), into an efficient rendering pipeline through a transparent, parametric model (\cref{sec:simulationmodel}) and the corresponding derivation of relevant parameters through a dedicated optical characterization of the sensor (\cref{sec:optical-characterization}). This approach is complementary to prior work in that it integrates optical measurement parameters to more accurately simulate LiDAR data. Depending on the amount of information available about the environment and target objects, and depending on measurement capabilities, approaches may be preferable that can operate exclusively on recorded scenario data. The presented model, in turn, provides analytical solutions for such phenomena for cases where dedicated laboratory measurements are feasible and a minimal domain gap is desired.


\subsection{Physically-based rendering and BRDFs}\label{sec:intro-pbr}

The model adopts the physically-based rendering (PBR) framework \cite{pharr2023physically}, a set of widely used and supported principles for optical image generation, particularly for modeling visual surface properties. The underlying surface model is the bidirectional reflectance distribution function (BRDF) over spherical angles $\spangle = (\phi, \theta)$, describing the relationship between incident and reflected light at a surface interaction, via
\begin{equation}
    \brdf(\spangle\subs{i}, \spangle\subs{o}) = \dee L\subs{o}(\spangle\subs{o}) / \dee E_i(\spangle\subs{i}),\label{eq:brdf}
\end{equation}
where $\spangle\subs{i}$ is the incident light direction, $\spangle\subs{o}$ is the reflected or outgoing light direction, $E\subs{i}$ is the incident light \emph{irradience}, and $L\subs{o}$ is the outgoing \emph{radiance}. The general framework enables two commonly used sets of units: photometric and radiometric. Photometric units are based on \emph{luminous flux} (as measured in Lumen, with Candela and Lux as derivable units), which in turn is based on the perception of the human eye. Radiometric units use raw physical quantities around \emph{power} (as measured in Watt). Both are interchangeable for the PBR framework, by which light emission intensities and surface interactions are combined to derive received / perceived intensities. For computer graphics, photometric units are generally preferred, as they relate to visual appearances. For LiDAR simulation, however, the response in the laser domain (usually infrared) and power properties of the scanner must be modeled. Hence, radiometric units apply for this use case, with Watt being by far the most common unit.

However, absolute physical units must be considered with caution for inferring simulation parameters, as these are strongly determined by sensor-intrinsic properties---both for echo intensity, which is not recorded directly but through the sensor's AD detection threshold, and the ratio between echo and ambient light intensity at which an echo detection is possible, which depends on the sensor's signal processing. Hence, these properties are determined in a data-driven way as well, by which intensities reduce to relative values.

A particular parametrization of such a BRDF was presented in \cite{burley2012physically} under the name of ``Disney principled BSDF'' (where the ``S'' generalizes ``reflectance'' to ``scattering''), describing the BRDF at each surface point through a set of intuitive parameters, each with plausible values within a common range of $[0, 1]$, such as diffuse, roughness, metalness, and surface normals. This concept sparked a family of related  models that have been used not only in renderers (raytracers) such as Mental Ray, Mitsuba, or Cycles, but also in shader-based\nobreak{
 /} rasterization-based game engines, such as Unreal, Unity, or OGRE3D. While generalizations exist for volumetric information of materials, e.g., refraction and subsurface scattering, the most common use cases are pure surface interactions.

\subsection{Blooming phenomena}\label{sec:intro-blooming}
The point cloud reported by a LiDAR sensor cannot perfectly match the scanned 3D scene. Apart from inevitable noise, point clouds exhibit artifacts, which stem from the measurement principle. One of these artifacts is blooming, which refers to the apparent widening of the contour of objects with a retro-reflective surface. 
Retro-reflection describes the property of optical surfaces that reflect incoming light back towards its origin, irrespective of the direction of incidence. The ideal BRDF of a retro-reflector approximates a Dirac distribution for $\spangle\subs{o} = \spangle\subs{i}$. The BRDF of real retro-reflectors used in road traffic has a peak reflectance $\brdf(\spangle\subs{i}, \spangle\subs{o})$ in the range of 100 to 1\,000, and a rapid drop-off for increased angles spanned by $\spangle\subs{i}$ and $\spangle\subs{o}$. Since the working principle of a LiDAR sensor is based on sampling back-reflected radiation, retro-reflecting surfaces always exhibit their peak reflectance from the perspective of the sensor. Compared to diffuse surfaces, whose BRDF cannot exceed $\pi^{-1}$, the retro-reflector is approximately three orders of magnitude brighter.

The LiDAR emits a divergent laser beam whose angle-resolved radiant intensity roughly resembles a peak that quickly declines.
The outline of any surface can bloom to a limited degree if a subsection of the beam intersecting the border causes enough reflection for the sensor to detect it. The sensor effectively integrates over all incoming radiation, attributing the ToF measurement to the center direction of the beam, which lies outside the target.
Because of the high reflectance of retro-reflectors, even minute radiant intensities in the periphery of the emitted pulse cause sufficient reflection to detect the surface, possibly dominating incoming reflections from diffuse surfaces hit by different regions of the pulse.
Since blooming can cause the contour to widen by several degrees, and due to the prevalence of retro-reflectors on roads and vehicles, blooming is a relevant aspect of the domain gap for automotive LiDAR use-cases. Most systems developed for assisted or automated driving rely on the absolute distance reference a LiDAR sensor provides, for orientation and classification of the surroundings. Thus, it is essential to test the robustness of these systems towards blooming artifacts.

\begin{figure}
    \centering
    \begin{overpic}[
    width=\linewidth]{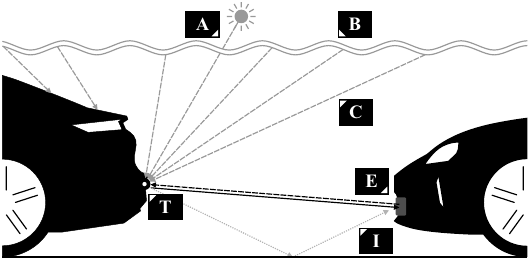}
    \put(50,15){\small $\emitter_\beam(\spangle)$}
    \put(50,7){\small $\reflected_\beam(\spangle)$}
    \end{overpic}
    \caption{Included and excluded aspects of the single-bounce model: LiDAR rays originate at the emitter/collector \textbf{E} with intensity $\emitter_\beam(\spangle)$, continue to a single target spot \textbf{T}, and return at the same angle with intensity $\reflected_\beam(\spangle)$. Stray light travels from a source (e.g., the sun \textbf{A}) to \textbf{T}, then to \textbf{E}. Virtual sources \textbf{C}, (e.g., atmospheric scattering \textbf{B}), can also be considered. However, indirect paths from/to \textbf{E}, such as multiple bounces over the road \textbf{I}, are \emph{not} modeled.}
    \label{fig:single-bounce}
\end{figure}

\section{Simulation Model}\label{sec:simulationmodel}

The following simulation model is implemented in the \octas{}\footnote{\href{https://www.octas.org}{octas.org}\quad\footnotemark[2]\href{https://www.ogre3d.org}{ogre3d.org}\quad\footnotemark[3]\href{https://www.cycles-renderer.org}{cycles-renderer.org}} simulation framework, which provides a modular architecture that enables, in particular, the exchange of different rendering engines---for the given use case, the plugins for the rasterization-based engine OGRE3D\footnotemark{} and for the raytracing engine Cycles\footnotemark{} are used, which generate consistent outputs based on the PBR framework.

\subsection{LiDAR model}

We establish a \emph{single-bounce model}  (cf. \cref{fig:single-bounce}), s.t. only a single geometric interaction between a light source and a surface is modeled before the light is received. Interactions that can be included in either the local surface (such as retroreflection or clearcoat) or the ambient light source (such as atmospheric lighting) may also be considered within the model. Generalizations towards multiple bounces, in particular reflections or refractions of LiDAR beams, are omitted here, as they considerably increase computational effort while effects will usually not be dominant (cf. \cref{fig:lidarsim-bounces}). In the following, we use the spherical angle $\spangle \in \spangleSpace$ and ranges $r \in \rangeSpace$.

We model a LiDAR as having a set $\Beams = \{\beam_1, \beam_2, ... \}$ of beams, which we use as indices for particular properties. These properties include the central directional angle per beam, $\beamangle_\beam$ in (sensor-fixed) spherical coordinates; a distribution of emitted intensities $\emitter_\beam : \spangleSpace \to [0, \infty)$; a resulting reflected intensity of light from the scene at different ranges $\reflected_\beam : \spangleSpace \times \rangeSpace \to [0, \infty)$; and an angular sensitivity of the collecting diode $\collector_\beam : \spangleSpace \to [0,1]$; leading to a signal $\signal_\beam : \rangeSpace \to [0, \infty)$ of received intensities over different ranges. These are related via
\newcommand\sphericaldelta{\dee\omega_\spangle}
\begin{equation}
    \textstyle \signal_b(r) = \iint\!\sphericaldelta\; \collector_b(\spangle)\;\reflected_b(\spangle,r)\;\;\text{with}\;\; \sphericaldelta = {\sin}\theta\,\dee\theta\,\dee\phi\;.
\end{equation}
We note that this equation does not explicitly include all of the aforementioned quantities: $\beamangle_\beam$ is implicitly contained in $\emitter_\beam$ but not required separately, and $\emitter_\beam$ certainly affects $\reflected_\beam$ through the individual BRDFs $\brdf$ of the surfaces in the scene, along with their geometries. The latter relationship ($\emitter$, $\reflected$, $\brdf$) is complex, but---fortunately---solved in a very mature way through PBR frameworks at various levels of detail. 
Unlike many applications for graphical rendering, which only require $\reflected_b(\spangle)$ (a 2D image), we require volumetric information $\reflected_b(\spangle,r)$. However, due to the single-bounce approach, we can assume to be in possession of $r(\spangle)$, a unique range at each spherical angle, at least for any target contributing $\reflected_b(\spangle,r) > 0$. Often this information is available, though not commonly demanded in the end product of rendering: Raytracing will typically compute depths directly, whereas rasterization-based rendering commonly requires the computation of a \emph{z buffer} to handle occlusions (cf. also \cref{sec:rendering}), which stores pixel $z$ coordinates (perpendicular to the image plane), leading to
\begin{equation}
    \textstyle \signal_\beam(r) = \iint\!\sphericaldelta\; \collector_\beam(\spangle)\;\reflected_\beam(\spangle)\;\delta(r - r(\spangle)),
\end{equation}
where $\delta$ is the Dirac delta, meaning that for $\eta_\beam(r)$ only intensities are accumulated which are at range $r$. For brevity, we omit that ambient light (e.g., sunlight) is computed identically except for substituting the emitted light by any ambient light sources, acting as the intensity of the noise floor against which echoes must be detected. 
This statement provides a first complete model to establish received signal intensities $\signal_\beam$ for a given beam $\beam$ based on a rendering of scene reflections $\reflected_\beam(\spangle)$ and corresponding depths $r(\spangle)$, and the collector sensitivity $\collector_\beam$. Yet, computational effort is high when more than a single beam is considered: Each beam requires the rendering of one scene, which is typically prohibitive for modern LiDAR systems with large, dense point clouds. Hence, simplifications are required that will be discussed in the following.

\subsection{Simplifications}

Since the scanner's intrinsic geometry is usually very small compared to the scale of the scene, all beams can be considered as originating from a single point coinciding with the collector. Because of this, overlapping beams share the same optical path for a surface point of the scene geometry. In \eqref{eq:brdf}, the considered $\spangle\subs{i}$ and $\spangle\subs{o}$, as well as the resulting $\brdf$ will reappear for multiple beams. The same is true for the depth $r(\spangle)$. Since typically the initialization of a new rendering $\reflected_b(\spangle)$ is a particularly costly operation, this motivates the combination of multiple beams in a single rendering call.

We define $\reflectedN(\spangle)$ as the reflected intensities for a uniform, isotropic emission source (a point light in the scene), s.t.
\begin{equation}
    \textstyle \signal_\beam(r) = \iint\!\sphericaldelta\; 
    \collector_\beam(\spangle)\;\underbrace{\emitter_\beam(\spangle)
    \reflectedN(\spangle)}_{\reflected_\beam(\spangle)}\;\delta(r - r(\spangle))\;,
\end{equation}
utilizing that in the single-bounce model any rays are directly between sensor and surface along $\spangle$, and no secondary reflections must be considered. Further assuming that $\collector_\beam$ and $\emitter_\beam$ only depend on $\beam$ through an angular offset by $\beamangle$, we can define ``centered'' kernels $\collector$, $\emitter$ to simplify to
\begin{equation}
    \textstyle \signal_\beam(r) = \iint\!\sphericaldelta\; 
    (\collector\cdot\emitter)(\spangle-\beamangle_\beam)\;
    \reflectedN(\spangle)\;\delta(r - r(\spangle))\;.
\end{equation}
This resembles the convolution or cross-correlation $(\collector\cdot\emitter)*\reflectedN$, with a significant difference: The $\delta$ effects that individual depths must be considered independently, s.t. a single global convolution (as for camera-based blooming) is not applicable.

\begin{figure}
    \begin{subfigure}{\columnwidth}
    \begin{overpic}[
    width=\linewidth]{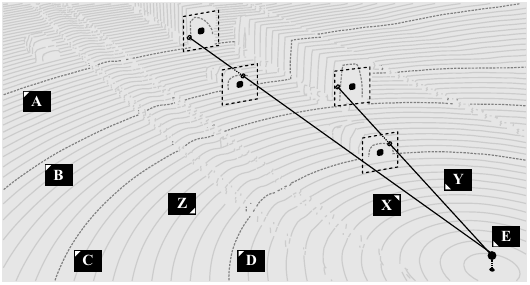}
    \end{overpic}
    \caption{Bird's eye view of the scene with \emph{iso ranges} (not scanning layers)}
    \end{subfigure}\\[4pt]

    \begin{subfigure}{\columnwidth}
    \begin{overpic}[
    width=\linewidth]{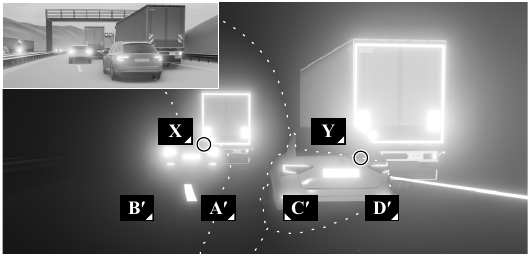}
    \end{overpic}
    \caption{Exemplary rendering of visual image and LiDAR echo at dominant ranges} 
    \end{subfigure}\\[4pt]

    \caption{Principle of range-separated blooming for first dominant echo. Lines \textbf{Z} in the bird's eye view indicate iso ranges. Signals along lines of sight (\textbf{X}, \textbf{Y}) do not mix additively across different ranges. Instead, the strongest echo is propagated, forming regions (\textbf{A$'$}--\textbf{D$'$}) in which retroreflections of one of the four front vehicles, and hence their ranges \textbf{A}--\textbf{D}, dominate. \textbf{X} and \textbf{Y} are located at spatial angles at which the echoes of the respective car and truck have matching intensity. Optical effects are exaggerated for visibility, and only the range dominance among \textbf{A}--\textbf{D} is compared.}
    \label{fig:blooming}

\end{figure}

\subsection{Algorithmic solutions}

Based on these simplifications, two distinct algorithms are outlined, presented in \cref{alg:beam-iteration} (beam iteration) and \cref{alg:range-stacking} (range stacking) respectively. The presented algorithms focus on exposing the key principles in the most concise and clear form, leading to differences in processing aspects and interfaces. It should be noted that both algorithms essentially work on the same types of input and can be used (with additional pre- and post-processing steps) interchangeably. When using identical parameters and outputs, results of either algorithm should correspond to within floating point arithmetic limits.

A common post-processing step (not elaborated in detail here) will entail the computation of echo pulse widths (EPW), which serves as a surrogate for echo intensity for receivers that contain only binary thresholds $T$ at their AD conversion. 
Emitted pulses have a non-vanishing length in time (and thus range), which we denote $\varpi(r)$. Hence, such scanners will receive $\hat\signal = \varpi*\signal$ and record intervals $[r_0, r_1]$ s.t. $\forall r \in [r_0, r_1] : \hat\signal > T \wedge \nexists \varepsilon_0, \varepsilon_1 > 0 : \forall r \in [r_0-\varepsilon_0, r_1+\varepsilon_1] : \hat\signal > T$, i.e., the intervals during which $\hat\signal$ exceeds $T$. For perpendicular targets, $\signal(r) \approx a \cdot \delta(r-r\subs{target})$, where $a$ is the reflected amplitude and $\delta$ is the Dirac delta, and hence $\hat\signal(r) \approx a \cdot \varpi(r-r\subs{target})$. Now when measuring the range interval over which $a \cdot \varpi(r-r\subs{target}) > T$, its width will usually (e.g., for Gaussian-shaped $\varpi$) scale with $a$. Hence, the EPW $r_1-r_0$ correlates with $a$, most strongly so for perpendicular targets. For flat incidence angles, EPW will be dominated by the actual geometric range interval over which the object reflects the beam.

Furthermore, for simplicity, the descriptions do not address that raw rendered images will typically follow a pinhole camera projection rather than spherical coordinates, in which kernels $\collector\emitter$ would be inhomogeneous for homogeneous beams in spherical coordinates. Different solutions to this exist, from adapting the kernel (only for \cref{alg:beam-iteration}), over the reprojection of rendered images, to the subdivision of wider fields of view into smaller, narrower renderings.

\subsubsection{Beam iteration}\label{sec:beam-iteration}

\begin{algorithm}[t]
\small\setstretch{\algostretch}
\Input{$\reflectedN[U][V]$: rendered intensities\newline
$\varrho[U][V]$: rendered ranges \newline
$\nu[U][V]$: rendered normals \newline
$\collector\emitter[2M+1][2N+1]$: centered kernel\newline
$\reflected\subs{min}$: threshold intensity\newline
$\Delta r$: range resolution\newline
$\varrho\subs{nearest}$: maximum range\newline
$\Beams$: Set of beams
}
\KwResult{%
$\signal[|\Beams|][R]$: received intensities over ranges
}

$\signal \gets (0)_{|\Beams|\times R}$\;

\For{$\beam \in \Beams$}
{
    $\beamangle \gets \text{beamAngle}(\beam)$\;
    $\hat\beamangle \gets \text{roundToPixels}(\beamangle)$\;

    \For{$(m,n) \in \{-M, ..., M-1, M\} \times \{-N, ..., N-1, N\}$}
    {

				$[u\subs{offs}, v\subs{offs}] \gets [\hat\beamangle\subs{u} + m, \hat\beamangle\subs{v} + n]$\;
        $[r_0, r_1] \gets \displaystyle\text{pixelRangeInterval}(u\subs{offs}, v\subs{offs}, \varrho, \nu)$\;
			  \For{$r\subs{offs} \in \{r_0, r_0 + \Delta r, ..., r_1\}$}
				{
            $\signal[b][r\subs{offs}] \gets \reflectedN[u\subs{offs}][v\subs{offs}] \cdot \collector\emitter[m][n]$\;
				}
    }
}
\setstretch{1.0}
\caption{Outline of the beam iteration algorithm as described in \cref{sec:beam-iteration}. See there also for a description of pixelRangeInterval().  Time complexity is $\mathcal O(M\cdot N \cdot |\Beams|)$ for single echoes, and $\ldots\cdot (\Delta r)^{-1}$ for multiple echoes.\vspace{-15pt}}\label{alg:beam-iteration}
\end{algorithm}

This is the more direct and comprehensive solution, described in \cref{alg:beam-iteration}. Beams are processed individually based on precomputed images, and for each beam, an array of echo signals over range (or time, respectively) is maintained. This signal array is composed of the contributing intensities over the kernel $\collector\emitter$, where each rendered pixel contributes uniformly to a range interval according to a function pixelRangeInterval(), which yields an estimate for the range interval contained within the pixel by extrapolating from pixel range and surface normal. To obtain a single echo per beam (comparable to \cref{alg:range-stacking}), the per-beam intensities can be searched for strongest / nearest / longest echoes. For the former two, only maxima must be stored, and hence only space of $\mathcal{O}(|B|)$ must be allocated rather than $\mathcal{O}(|B|\cdot R)$, and the post-processing step is avoided.

\subsubsection{Range stacking}\label{sec:range-stacking}

\begin{algorithm}[t]
\small\setstretch{\algostretch}
\Input{$\reflectedN[U][V]$: rendered intensities\newline
$\varrho[U][V]$: rendered ranges \newline
$\collector\emitter[2M+1][2N+1]$: centered kernel\newline
$\reflected\subs{min}$: threshold intensity\newline
$\Delta r$: range resolution\newline
$r\subs{max}$: maximum range
}
\KwResult{%
$S\subs{nearest}[U][V]$: dominant range indices\newline
$\signal\subs{nearest}[U][V]$: dominant intensities
}
$s\subs{max} = \lfloor r\subs{max} / \Delta r \rfloor$\;
$\bar{\varrho}\gets \lfloor\, \varrho / \Delta r\, \rfloor$\;
$\signal\subs{nearest} \gets (0)_{U\times V}$\;
$S\subs{nearest} \gets (s\subs{max})_{U\times V}$\;
\For{$s \gets s\subs{max}$ \KwTo $1$}
{
    $\reflectedN\subs{slice} \gets (0)_{U\times V}$\;
    \For{$(u,v) \in \{1, ..., U\} \times \{1, ..., V\}$}
    {
        \lIf{$s = \bar{\varrho}[u][v]$}
        {%
            $\reflectedN\subs{slice}[u][v] \gets \reflectedN[u][v]$%
        }
    }

    $\signal\subs{slice} \gets \reflectedN\subs{slice} * \collector\emitter$\;\label{algln:range-stacking-convolve}
    \For{$(u,v) \in \{1, ..., U\} \times \{1, ..., V\}$}
    {
        \If{$\signal\subs{slice}[u][v] \geq \reflected\subs{min}$}
        {
            $\signal\subs{nearest}[u][v] \gets \signal\subs{slice}[u][v]$\;
            $S\subs{nearest}[u][v] \gets s$\;
        }
    }
}
\setstretch{1.0}
\caption{Outline of the range stacking algorithm for single closest echo above threshold. Note that the convolution in ln.~\ref{algln:range-stacking-convolve} can be optimized for separable kernels, to yield a time complexity of $\mathcal O((\Delta r)^{-1} \cdot U\cdot V)$, and $\ldots + |\Beams|$ for sampling actual beam ranges in post-processing.\vspace{-15pt}}\label{alg:range-stacking}
\end{algorithm}

For high numbers of very dense beams, relatively narrow fields of view, and moderate range resolution, it can be beneficial to replace the iteration over the kernel cells in $\collector\emitter$ by actual convolutions---however, as initially remarked, such convolutions must handle ranges independently.

In this range stacking approach, the rendered images are split into discrete slices according to the depth resolution of the sensor (or an approximation thereof), where typically (as in \cref{alg:range-stacking} only one such slice is kept in memory at a time.

Intensities within the same slice are then convolved, and can iteratively be combined with other slices to propagate, e.g., the strongest echo from far to near ranges. While this will typically involve computing a much denser pattern of beams (namely at pixel level as in \cref{fig:blooming}) and prohibit the storage of individual echo arrays over range, the benefit lies in replacing the manually implemented pseudo-convolution by a set of true convolutions. These can leverage various optimizations, such as the parallel computation on dedicated hardware (namely GPUs), reducing complexity for (near-) separable kernels (where $\collector\emitter \approx H * V$ with $H^\top{\in}\,\mathbb R^{M}, V \,{\in}\,\mathbb R^{N}$, s.t. $ \reflectedN * \collector\emitter \approx  ((\reflectedN * H) * V)$, cf. also \cref{sec:optical-angle-sensitivity}) from $\mathcal O(N\cdot M)$ to $\mathcal O(N + M)$, or even box kernels via differences in integral images. These optimizations are particularly advantageous when the overlap between beam kernels is significant s.t. avoiding their repeated computation provides a notable gain.




\subsection{Practical rendering considerations}\label{sec:rendering}

The model can be used with raytracing as with rasterization-based rendering within the PBR framework, where the latter is favorable for many use cases due to its faster rendering speed. The choice of a rendering engine may introduce limitations that require addressing. The following sections give a brief overview of the main relevant effects and solution strategies. 

\subsubsection{z buffer resolution}

When used as the source for depth information, the depth resolution of the z buffer must be considered. As it is designed for visual rendering purposes, the resolution is not distributed homogeneously over the z depth (which, in turn, would not correspond to homogeneous range resolution, anyway), but reciprocally between the near and far clipping planes, s.t. close depths are resolved more accurately than distant ones. 
Accordingly, the resolution of the depth buffer directly affects the range resolution particularly at the far ranges, and particularly through the choice of the \emph{near} clipping plane, since most resolution units are allocated there. Common z buffer depths for realtime rendering (as in OpenGL, for example) are 16, 24, or 32 bits. \cref{fig:normals-clipping} shows errors arising based on the variation of the near and far plane.

\subsubsection{Rasterization}

The most commonly found simplification is rasterization: Images are sampled in a regular pixel grid, which typically does not align with the scan pattern of a LiDAR system. This is a pivotal aspect in realtime GPU rendering on graphics cards, hence referred to as rasterization-based rendering. The acceleration structure of parallel computing can be leveraged optimally when perspective projections are the principal geometric transformations. Any nonlinear mappings considerably increase the computational effort and do not generally benefit from optimized existing solutions. Raytracing-based rendering in turn does not generally benefit as significantly from regular, grid-based sampling; however, it is the most commonly used interface for graphics rendering. Hence, some renderers such as Mitsuba directly support arbitrary ray directions, whereas others such as Cycles use raster images as their primary interface. 

\begin{figure}
\includegraphics{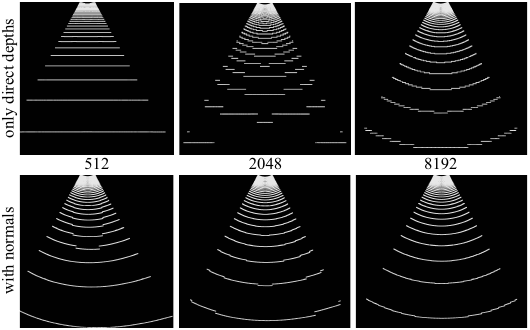}

\caption{Comparison of increased resolution vs. the correction (through 10-bit normals) for reducing angular errors. It can be seen that even at high resolutions of $8192$ pixels wide, step artifacts remain distinct at far ranges. In contrast, when correcting via normals, residual steps exclusively arise from lack of precision in normals, if applicable. Increased resolution in this case also serves to eliminate extrapolation errors from erroneous normals.}
\label{fig:normals-errors}
\end{figure}

Errors introduced by rasterization are fundamentally angular errors, as information is sampled at discrete $\spangle$ which do not generally coincide with the ray directions of the scan pattern. In practice, these angular errors manifest as range errors, since ray directions are assumed to be accurate while ranges are read from the closest pixel coordinates, as seen in \cref{fig:normals-errors}.

Interpolation of ranges between pixels is typically out of the question in the given application: Changes in range across the sensor's field of view commonly stem from distinct objects at different depths, where no continuous transition can be assumed (as opposed to, for example, terrain models). Interpolation will thus generate intermediate ``ghost points'' between targets, whereas LiDAR systems will typically only yield ranges at which actual targets exist in the scene. The scale of the resulting errors is limited only by the maximum distance of the objects in the scene.

%

\begin{figure}
\includegraphics{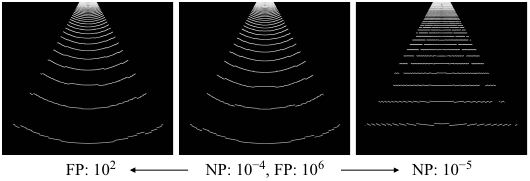}

\caption{Effect of clipping planes and thus z buffer resolution with normals correction. A near plane of $10^{-4}$ and a far plane of $10^6$ provide fair results. Pulling the far plane to $10^2$ provides little improvement. Pushing the near plane up to $10^{-5}$, in turn, yields degenerate, discrete depth levels.}
\label{fig:normals-clipping}
\end{figure}

To correct angular errors without the need for interpolation between different raster pixels, a linear extrapolation of a given range based on surface normals can be performed according to
$r_{\text{corr}}=\bm{p}\transp \bm{n}/(\bm{v}\transp \bm{n})\cdot r$
where $\bm p$ is the direction vector towards the pixel center, $\bm v$ is the true direction of the LiDAR ray, $\bm n$ is the surface normal, and $r$ is the range at the pixel center.

It must be noted that normal precision may be limited in the rendering process, as these are, in realtime graphics applications, commonly used for effects where accuracy is less important (cf. \cref{fig:normals-clipping} for the impact of z buffer resolution). \cref{fig:normals-clipping,fig:normals-errors} show 10 bit \emph{normals}, whereas \emph{z buffer} depths of 16 or 32 bits are common. Hence, for well-chosen clipping planes, errors from normals are the more likely error source.

\section{Optical Characterization}\label{sec:optical-characterization}
Both the beam intensity $\emitter(\spangle)$ and the sensitivity $\collector(\spangle)$ are measured with directional dependence relative to the main beam direction. If both $\emitter$ and $\collector$ are assumed homogeneous over the sensor's scan pattern, only a single beam direction needs to be characterized and generalized for all measurement directions. Any deviations can be subsequently parameterized as stretched or rotated versions of the reference pulse.

\subsection{Angle-resolved beam intensity}
The measurement method used to characterize $\emitter(\spangle)$ is borrowed from far field photogoniometry, which deals with the measurement of the angularly resolved luminous intensity of light sources. The measurement setup consists of a detector and a goniometer, which is used to precisely rotate the light source around two rotation axes towards the detector.

The setup in question employs an imaging detector system consisting of a diffuse screen, which is illuminated by the source, and a spectrally and geometrically calibrated camera pointed at the screen. This setup allows measuring the luminous intensity for many directions using a single image with the angular resolution only limited by the camera's resolution. For LiDAR sensors, this setup can be used to measure relative radiant intensities in the near-infrared, where the sensor of the camera is still sensitive. For beam characterization, the camera is used in combination with a long-pass filter, which is transparent in the infrared and blocks visible light.

Measurement of beam intensity poses requirements for the scan pattern. The spacing of the measurement directions needs to be sufficiently large so that they do not overlap. A more general requirement is a constant scan pattern, which allows the camera to integrate over multiple realizations of the same beam direction during the imaging process. The repeatability of a LiDAR sensor is normally accurate enough, so this integration can be reasonably performed. The Blickfeld Cube1 LiDAR exposes many settings to configure the scan pattern and is therefore used for the following exemplary measurements.

\cref{fig:blickfeld-dots} shows a goniometer measurement for a section of the scan pattern of the Blickfeld Cube1. The obtained beam intensity can be used for simulations, but its intensity resolution is not sufficient to simulate blooming artifacts. Since blooming is caused by the very dim outer edges of the pulse shape, a measurement with a high dynamic range is required to resolve the beam intensities across at least six orders of magnitude. To increase the sensitivity, the measurement setup is adapted to use a retro-reflective screen and a camera positioned close to the LiDAR sensor. The screen is constructed from micro prism-based retro-reflective contour marking tape for trucks and trailers. Since the angle spanned by the LiDAR source and the camera is small enough, the received signal is amplified drastically by the reflectance of the retro-reflector. 
The measurement result for an isolated beam is shown in \cref{fig:blickfeld-singledot}. 
The effective extent of the pulse measures approx. 3°---six times larger than the core pulse shape, providing the optical reason for blooming.

\newcommand\ocfigscale{0.73}
\newcommand\ocfigvoffset{-3mm}
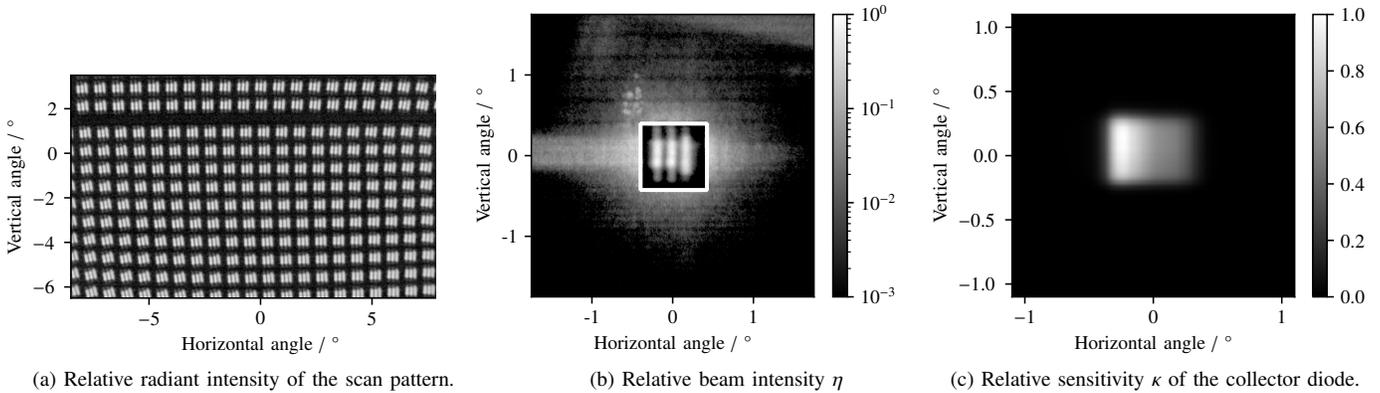
\begin{figure*}
    \begin{subfigure}[b]{0.35\linewidth}
        \centering
        \scalebox{\ocfigscale}{\hspace{-0.7cm}\input{images/LVK.pgf}}\vspace{\ocfigvoffset}
        \caption{Relative radiant intensity of the scan pattern.}\label{fig:blickfeld-dots}
    \end{subfigure}
    \hfill
    \begin{subfigure}[b]{0.3\linewidth}
    \centering
    \scalebox{\ocfigscale}{\hspace{-1.5cm}\input{images/pulse.pgf}}\vspace{\ocfigvoffset}
    \caption{Relative beam intensity $\emitter$ }\label{fig:blickfeld-singledot}
    \end{subfigure}
    \hfill
    \begin{subfigure}[b]{0.3\linewidth}
    \centering
    \scalebox{\ocfigscale}{\hspace{-1cm}\input{images/sensitivity.pgf}}\vspace{\ocfigvoffset}
    \caption{Relative sensitivity $\collector$ of the collector diode.}\label{fig:blickfeld-collector}
    \end{subfigure}
    \vspace{3mm}
    \caption{Results of the optical characterization of the Blickfeld Cube 1.  Each group of three spots corresponds to one beam direction, as the sensor employs three laser diodes to construct an approximately square pulse shape. Highly-resolved beam intensity in (b) is measured with retro-reflective screen. The inner section of the pulse shape marked with a white rectangle is scaled down by a factor of 1\,000 to visualize the entire pulse within the same range.    
    While offering the highest grade of retro reflectance, the prism optics introduce a higher dependency on the incidence direction and a speckled appearance. Dark stripes result from the horizontal application of the tape. A more uniform retro-reflective surface based on glass beads could be more suitable. Relative sensitivity in (c) over input angles is composed of separate horizontal and vertical measurements via $H * V$.}
\end{figure*}

\subsection{Angle-resolved collector sensitivity}\label{sec:optical-angle-sensitivity}
The sensitivity of the receiving optics generally decreases for radiation coming from directions that deviate from the beam direction $\beamangle$. While not strictly necessary for measuring the collector's sensitivity, it is beneficial if the LiDAR sensor reports some measure of intensity for the incoming radiation. The output of the Cube 1 includes a measure for ambient (passive) intensity for every measurement direction, which scales linearly with incoming radiation. The input sensitivity was measured by aiming a single beam of the LiDAR sensor at a distant halogen light source using the goniometer. This light source, in turn, creates a constant irradiance on the sensor. The change in the reported ambient intensity is then tracked while the sensor is turned away from the main beam direction in 0.01° steps. This is done for both the horizontal and vertical direction, resulting in two input sensitivity slices, which are treated as two 1D components of a separable 2D filter kernel. The combined input sensitivity is displayed in \cref{fig:blickfeld-collector}.


\subsection{Combined optical domain representation}
The multiplication of the emitter radiant intensity $\emitter$ and the collector sensitivity $\collector$ is employed in the LiDAR model as an effective beam intensity kernel. As the measurements for the emitter radiant intensity $\emitter$ and collector sensitivity $\collector$ are in relative quantities, the simulated signal $\signal$ would not allow conclusions about whether it would trigger the sensor. To set these relative values into relation, the detection threshold was determined using the same measurement setup and a dark diffuse target. From this measurement, an equivalent detection threshold signal is deduced. 


\begin{figure*}[t]
    \includegraphics{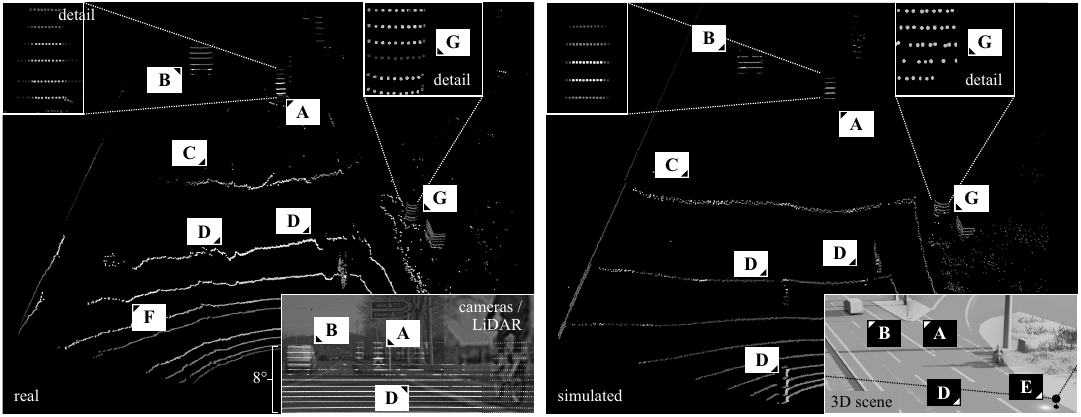}
    \caption{Comparison of consistently modeled objects between the real point cloud from Valeo Scala Gen. 2 and simulated point cloud (emitter located at \textbf{E}), points are colored by echo pulse width. \textbf{A} shows the blooming of the retroreflective sign, \textbf{B} is a leading van (whose respective intensities are underestimated in the simulation). The increase of the echo pulse width at flat incident angles \textbf{C} is qualitatively represented, whereas the unevenness of the road surface is underrepresented in the simulation, leading to weaker distortions at the ground \textbf{F}. The shape distortions in the simulated data \textbf{D} exclusively stem from blooming at the retroreflective road markings, which are found in the real world data in a comparable way. Furthermore it is seen at the roadside pole \textbf{G} that not all layers share the same intensity in the actual Scala sensor due to the diode boundaries. The simulated data, however, assumes constant pulse intensity.}
    \label{fig:lidarsim-example1}
\end{figure*}

\section{Real-World Application}

As the model is calibrated against the laboratory measurements through defined targets and goniometer, the strong match between the resulting real and synthetic point clouds provides limited insight into the validity of the model. To demonstrate the pipeline in whole and independently, the simulation is applied to real-world data recorded by a Scala Gen. 2 mounted to a research vehicle. In this use case, the limitation is that world properties (geometries and BRDFs) are only approximated in the simulation, enabling mainly qualitative rather than quantitative comparisons. Yet, the presence or absence of effects provides insight into model limitations.

An exemplary frame is shown and discussed in \cref{fig:lidarsim-example1}. It can be seen that blooming effects from the retroreflective sign \textbf{A} show good correspondence, while the kernel shapes differ in detail. Secondary blooming effects, namely the deformation of ranges near retroreflective road markings \textbf{D} also arise similarly in the synthetic and real data. Small road surface irregularities are not modeled in the simulation, yet they contribute considerable range-based errors at flat incidence angles \textbf{F}. Additionally, the idealized assumption that all beams share the same intensity does not hold for the Scala 2 sensor \textbf{G} due to emitter diode characteristics. It should be noted that these can be included in the proposed simulation models, but require additional laboratory measurements to calibrate.

The simulated frame contains points from 16\,384 beams (more than the true Scala Gen. 2 emits), sampled from a 4096~px wide rendered image by 68~px wide beam kernels. Rendering $\collector$ through OGRE3D takes a negligible duration of under 0.01$\,$s on an NVIDIA RTX 4000 Ada; whereas the computation of $\signal$, including beam diameters, presents the relevant performance bottleneck. Both algorithms are run on an Intel Core i7-13800H @ 2\,500 MHz with 14 cores; \cref{alg:beam-iteration} was implemented for purely sequential processing, yielding 9.5\,s of processing time per frame. \cref{alg:range-stacking} in contrast takes on average 4.3\,s for 1\,200 range bins (corresponding to $s\subs{max}$) at the same resolution, utilizing OpenCV convolutions through parallel processing utilizing the full CPU. Skipping depths with intensities entirely below a threshold sufficient to cause blooming can, for example, decrease computation time to 0.8\,s for only 180 remaining depth slices. 

\begin{figure}[t]
    \includegraphics{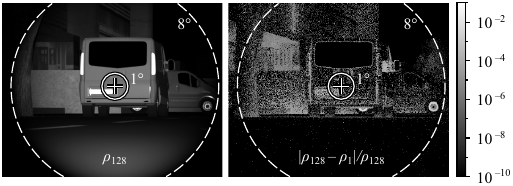}
    \caption{Deviations between the proposed single-bounce intensities for $\reflected_{1}$ and a full computation of 128 bounces $\reflected_{128}$ (each 128 samples/px) using the Cycles raytracer for a single beam with an extreme radius of 8°. Intensity differences are around $10^{-2}$ on the rough steel wheels, mostly however below $10^{-4}$.}
    \label{fig:lidarsim-bounces}
\end{figure}

Considering non-vanishing beam diameters is hence, despite the discussed optimizations, a very costly task. Options for parallel speedup outlined for the algorithms have thus far only been exploited in a limited way, but even then, the computation will typically be beyond real-time applications such as hardware in the loop (HiL). However, the proposed model and its algorithmic solutions enable synthesizing training and testing datasets including complex blooming artifacts and physically based surface interactions. \cref{fig:lidarsim-bounces} shows the practical effects of the single-bounce assumption on result intensities, indicating that this model assumption does not severely impact result quality, while providing substantial speedup.

\section{Conclusion and Outlook}

We have presented a model for the physically-based simulation of automotive LiDAR systems with a focus on the less considered aspects of beam geometries and surface interactions through PBR, such that model parameters can be obtained through laboratory measurements of the LiDAR systems rather than from sensor output data under real-world operations. The presented model enables the simulation of blooming and light interactions at different levels of detail via systematic simplifications and adaptations to established image rendering principles. It was shown that the model parameters can be extracted systematically for ToF LiDAR systems to yield simulations that reproduce known artifacts in LiDAR systems for automotive environments, in particular complex relationships arising without explicit modeling. Relevant computational and geometric considerations are provided to support practical applications in common rendering pipelines.


The presented model and its evaluation have several limitations that are left for future work. The model, as presented, requires a considerably more comprehensive demonstration on different LiDAR systems and the evaluation against heterogeneous use cases, in comparison with existing models. Primarily, this will include the generation of synthetic datasets to evaluate the efficacy of the model for creating testing and training data for AI / ML applications, quantifying the domain gap, as demonstrated in  \cite{manivasagam2023towards}. A second prerequisite for quantitative comparisons in complex environments is a means of deriving world object geometry and surface parameters with sufficient detail. A fundamental concern in simulations, particularly analytic models, is computational effort. While different levels of detail, simplifications, and two algorithmic solutions are discussed, options for leveraging parallel processing are not discussed in detail. Neither variant is realtime-capable on regular hardware, requiring several seconds per frame to compute. Hence, additional accelerations will be required to enable the approach for HiL testing, as opposed to dataset generation. Finally, several aspects of the optical characterization are not addressed herein, especially kernel derivation for scanners that do not provide single beam emissions and ambient (passive) light reception, and the relationship between AD threshold vs. ambient noise, which remain for future work.


\bibliographystyle{IEEEtran}
\bibliography{IEEEabrv,bibliography}

\end{document}

%% file: include/includes.tex
\usepackage{amsmath}
\usepackage{amssymb}

\usepackage[table]{xcolor}

\usepackage{tikz}
\usepackage[linesnumbered,vlined]{algorithm2e}
\usepackage{array}
\usepackage{framed, color}
\usepackage[normalem]{ulem}
\usepackage{cancel}
\usepackage{overpic}
\usepackage[footnotesize]{caption}
\usepackage[font+=footnotesize]{subcaption}
\usepackage[nodisplayskipstretch]{setspace}
\usepackage{txfonts}
\let\mathbb=\varmathbb

\usepackage{url}
\usepackage{upgreek}
\usepackage{float}
\usepackage{booktabs}
\usepackage{courier}
\newcommand{\href}[2]{#2}

\usepackage{nicefrac}
\usepackage{xcolor}
\usepackage{mdframed}
\usepackage{cleveref}
\usepackage{bm}

\usepackage{multirow}

%% file: include/setup.tex
\DeclareMathAlphabet{\mathbb}{U}{msb}{m}{n}

\newcommand{\dee}{\mathrm{d}}

\newcommand{\subs}[1]{_{\textrm{\upshape{#1}}}}

\newcommand{\transp}{^{\top}}

\crefname{figure}{Fig.}{Figs.}
\Crefname{figure}{Fig.}{Figs.}

\crefname{table}{Tab.}{Tabs.}
\Crefname{table}{Tab.}{Tabs.}

\crefname{equation}{Eq.}{Eqs.}
\Crefname{equation}{Eq.}{Eqs.}

\crefname{section}{Sec.}{Secs.}
\Crefname{section}{Sec.}{Secs.}

\crefname{subsection}{Sec.}{Secs.}
\Crefname{subsection}{Sec.}{Secs.}

\crefname{algorithm}{Alg.}{Algs.}
\Crefname{algorithm}{Alg.}{Algs.}

\newcommand\algostretch{1.0}

\SetKwComment{Comment}{/* }{ */}
\SetAlgorithmName{Alg.}{Alg.}{List of Algorithms}
\SetAlCapNameFnt{\footnotesize}
\SetAlCapFnt{\normalfont\footnotesize}
\SetKwInput{Input}{Input}

\newcommand\registered{$^\text{\textregistered}$}
\newcommand\octas{OCTAS\registered}

\newcommand\Beams{\mathcal B}
\newcommand\beam{b}
\newcommand\beamangle{\Psi}
\newcommand\emitter{\eta}
\newcommand\collector{\kappa}
\newcommand\reflected{\rho}
\newcommand\reflectedN{\hat{\reflected}}
\newcommand\signal{\xi}
\newcommand\brdf{\beta}
\newcommand\spangle{\Phi}
\newcommand\spangleSpace{\mathcal{A}}
\newcommand\rangeSpace{\mathcal{R}}

%% file: images/LVK.pgf
\providecommand{\mathdefault}[1]{#1}
\begingroup%
\makeatletter%
\begin{pgfpicture}%
\pgfpathrectangle{\pgfpointorigin}{\pgfqpoint{3.486924in}{2.440847in}}%
\pgfusepath{use as bounding box, clip}%
\begin{pgfscope}%
\pgfsetbuttcap%
\pgfsetmiterjoin%
\definecolor{currentfill}{rgb}{1.000000,1.000000,1.000000}%
\pgfsetfillcolor{currentfill}%
\pgfsetlinewidth{0.000000pt}%
\definecolor{currentstroke}{rgb}{1.000000,1.000000,1.000000}%
\pgfsetstrokecolor{currentstroke}%
\pgfsetdash{}{0pt}%
\pgfpathmoveto{\pgfqpoint{0.000000in}{0.000000in}}%
\pgfpathlineto{\pgfqpoint{3.486924in}{0.000000in}}%
\pgfpathlineto{\pgfqpoint{3.486924in}{2.440847in}}%
\pgfpathlineto{\pgfqpoint{0.000000in}{2.440847in}}%
\pgfpathlineto{\pgfqpoint{0.000000in}{0.000000in}}%
\pgfpathclose%
\pgfusepath{fill}%
\end{pgfscope}%
\begin{pgfscope}%
\pgfsetbuttcap%
\pgfsetmiterjoin%
\definecolor{currentfill}{rgb}{1.000000,1.000000,1.000000}%
\pgfsetfillcolor{currentfill}%
\pgfsetlinewidth{0.000000pt}%
\definecolor{currentstroke}{rgb}{0.000000,0.000000,0.000000}%
\pgfsetstrokecolor{currentstroke}%
\pgfsetstrokeopacity{0.000000}%
\pgfsetdash{}{0pt}%
\pgfpathmoveto{\pgfqpoint{0.619136in}{0.556118in}}%
\pgfpathlineto{\pgfqpoint{3.336924in}{0.556118in}}%
\pgfpathlineto{\pgfqpoint{3.336924in}{2.154816in}}%
\pgfpathlineto{\pgfqpoint{0.619136in}{2.154816in}}%
\pgfpathlineto{\pgfqpoint{0.619136in}{0.556118in}}%
\pgfpathclose%
\pgfusepath{fill}%
\end{pgfscope}%
\begin{pgfscope}%
\pgfpathrectangle{\pgfqpoint{0.619136in}{0.556118in}}{\pgfqpoint{2.717788in}{1.598699in}}%
\pgfusepath{clip}%
\pgfsys@transformcm{2.717788}{0.000000}{0.000000}{-1.598699}{0.619136in}{2.154816in}%
\pgftext[left,bottom]{\includegraphics[interpolate=false,width=1.000000in,height=1.000000in]{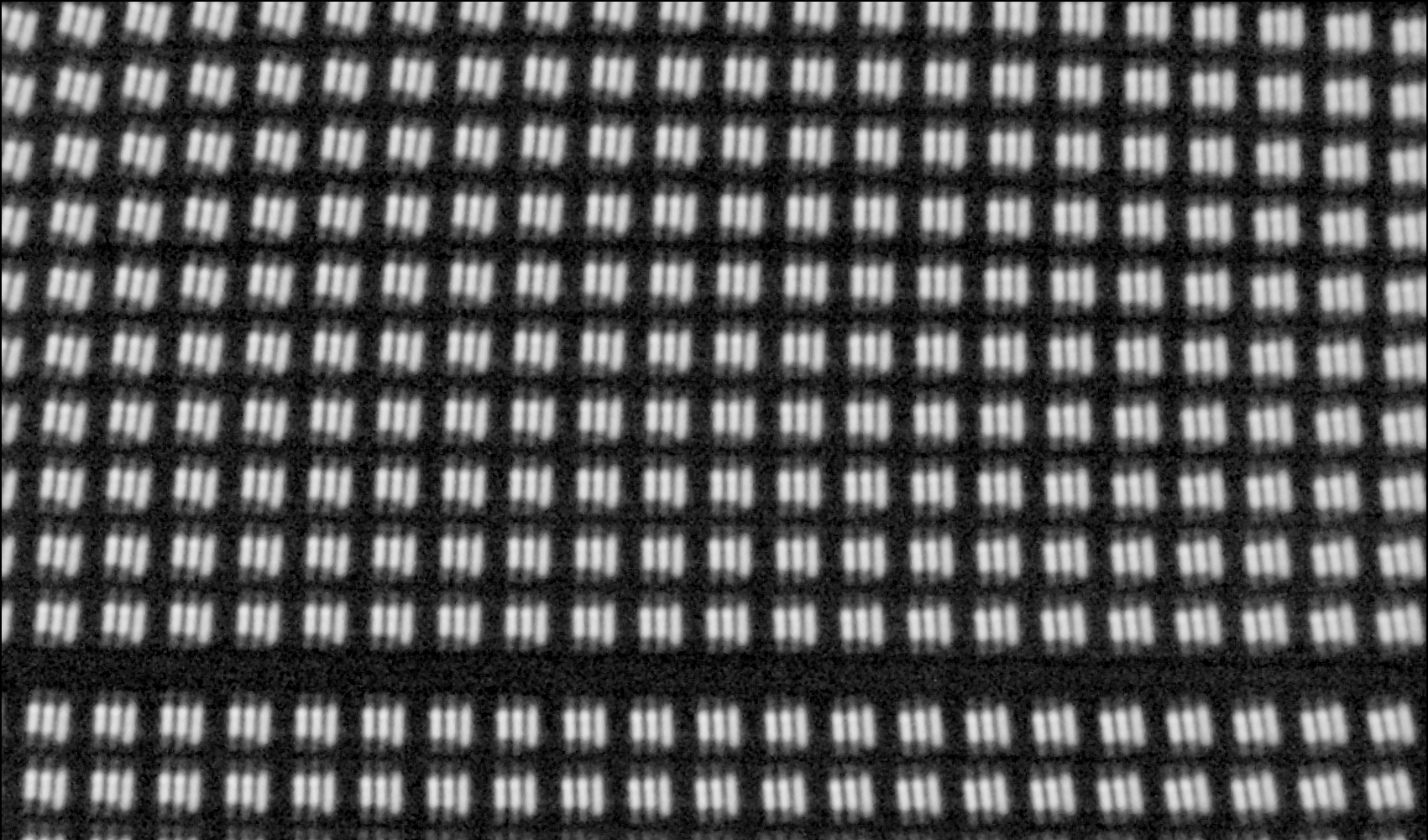}}%
\end{pgfscope}%
\begin{pgfscope}%
\pgfsetbuttcap%
\pgfsetroundjoin%
\definecolor{currentfill}{rgb}{0.000000,0.000000,0.000000}%
\pgfsetfillcolor{currentfill}%
\pgfsetlinewidth{0.803000pt}%
\definecolor{currentstroke}{rgb}{0.000000,0.000000,0.000000}%
\pgfsetstrokecolor{currentstroke}%
\pgfsetdash{}{0pt}%
\pgfsys@defobject{currentmarker}{\pgfqpoint{0.000000in}{-0.048611in}}{\pgfqpoint{0.000000in}{0.000000in}}{%
\pgfpathmoveto{\pgfqpoint{0.000000in}{0.000000in}}%
\pgfpathlineto{\pgfqpoint{0.000000in}{-0.048611in}}%
\pgfusepath{stroke,fill}%
}%
\begin{pgfscope}%
\pgfsys@transformshift{1.178681in}{0.556118in}%
\pgfsys@useobject{currentmarker}{}%
\end{pgfscope}%
\end{pgfscope}%
\begin{pgfscope}%
\definecolor{textcolor}{rgb}{0.000000,0.000000,0.000000}%
\pgfsetstrokecolor{textcolor}%
\pgfsetfillcolor{textcolor}%
\pgftext[x=1.178681in,y=0.458895in,,top]{\color{textcolor}{\rmfamily\fontsize{10.000000}{12.000000}\selectfont\catcode`\^=\active\def^{\ifmmode\sp\else\^{}\fi}\catcode`\%=\active\def
\end{pgfscope}%
\begin{pgfscope}%
\pgfsetbuttcap%
\pgfsetroundjoin%
\definecolor{currentfill}{rgb}{0.000000,0.000000,0.000000}%
\pgfsetfillcolor{currentfill}%
\pgfsetlinewidth{0.803000pt}%
\definecolor{currentstroke}{rgb}{0.000000,0.000000,0.000000}%
\pgfsetstrokecolor{currentstroke}%
\pgfsetdash{}{0pt}%
\pgfsys@defobject{currentmarker}{\pgfqpoint{0.000000in}{-0.048611in}}{\pgfqpoint{0.000000in}{0.000000in}}{%
\pgfpathmoveto{\pgfqpoint{0.000000in}{0.000000in}}%
\pgfpathlineto{\pgfqpoint{0.000000in}{-0.048611in}}%
\pgfusepath{stroke,fill}%
}%
\begin{pgfscope}%
\pgfsys@transformshift{1.978030in}{0.556118in}%
\pgfsys@useobject{currentmarker}{}%
\end{pgfscope}%
\end{pgfscope}%
\begin{pgfscope}%
\definecolor{textcolor}{rgb}{0.000000,0.000000,0.000000}%
\pgfsetstrokecolor{textcolor}%
\pgfsetfillcolor{textcolor}%
\pgftext[x=1.978030in,y=0.458895in,,top]{\color{textcolor}{\rmfamily\fontsize{10.000000}{12.000000}\selectfont\catcode`\^=\active\def^{\ifmmode\sp\else\^{}\fi}\catcode`\%=\active\def
\end{pgfscope}%
\begin{pgfscope}%
\pgfsetbuttcap%
\pgfsetroundjoin%
\definecolor{currentfill}{rgb}{0.000000,0.000000,0.000000}%
\pgfsetfillcolor{currentfill}%
\pgfsetlinewidth{0.803000pt}%
\definecolor{currentstroke}{rgb}{0.000000,0.000000,0.000000}%
\pgfsetstrokecolor{currentstroke}%
\pgfsetdash{}{0pt}%
\pgfsys@defobject{currentmarker}{\pgfqpoint{0.000000in}{-0.048611in}}{\pgfqpoint{0.000000in}{0.000000in}}{%
\pgfpathmoveto{\pgfqpoint{0.000000in}{0.000000in}}%
\pgfpathlineto{\pgfqpoint{0.000000in}{-0.048611in}}%
\pgfusepath{stroke,fill}%
}%
\begin{pgfscope}%
\pgfsys@transformshift{2.777379in}{0.556118in}%
\pgfsys@useobject{currentmarker}{}%
\end{pgfscope}%
\end{pgfscope}%
\begin{pgfscope}%
\definecolor{textcolor}{rgb}{0.000000,0.000000,0.000000}%
\pgfsetstrokecolor{textcolor}%
\pgfsetfillcolor{textcolor}%
\pgftext[x=2.777379in,y=0.458895in,,top]{\color{textcolor}{\rmfamily\fontsize{10.000000}{12.000000}\selectfont\catcode`\^=\active\def^{\ifmmode\sp\else\^{}\fi}\catcode`\%=\active\def
\end{pgfscope}%
\begin{pgfscope}%
\definecolor{textcolor}{rgb}{0.000000,0.000000,0.000000}%
\pgfsetstrokecolor{textcolor}%
\pgfsetfillcolor{textcolor}%
\pgftext[x=1.978030in,y=0.280007in,,top]{\color{textcolor}{\rmfamily\fontsize{10.000000}{12.000000}\selectfont\catcode`\^=\active\def^{\ifmmode\sp\else\^{}\fi}\catcode`\%=\active\def
\end{pgfscope}%
\begin{pgfscope}%
\pgfsetbuttcap%
\pgfsetroundjoin%
\definecolor{currentfill}{rgb}{0.000000,0.000000,0.000000}%
\pgfsetfillcolor{currentfill}%
\pgfsetlinewidth{0.803000pt}%
\definecolor{currentstroke}{rgb}{0.000000,0.000000,0.000000}%
\pgfsetstrokecolor{currentstroke}%
\pgfsetdash{}{0pt}%
\pgfsys@defobject{currentmarker}{\pgfqpoint{-0.048611in}{0.000000in}}{\pgfqpoint{-0.000000in}{0.000000in}}{%
\pgfpathmoveto{\pgfqpoint{-0.000000in}{0.000000in}}%
\pgfpathlineto{\pgfqpoint{-0.048611in}{0.000000in}}%
\pgfusepath{stroke,fill}%
}%
\begin{pgfscope}%
\pgfsys@transformshift{0.619136in}{0.636053in}%
\pgfsys@useobject{currentmarker}{}%
\end{pgfscope}%
\end{pgfscope}%
\begin{pgfscope}%
\definecolor{textcolor}{rgb}{0.000000,0.000000,0.000000}%
\pgfsetstrokecolor{textcolor}%
\pgfsetfillcolor{textcolor}%
\pgftext[x=0.344444in, y=0.587858in, left, base]{\color{textcolor}{\rmfamily\fontsize{10.000000}{12.000000}\selectfont\catcode`\^=\active\def^{\ifmmode\sp\else\^{}\fi}\catcode`\%=\active\def
\end{pgfscope}%
\begin{pgfscope}%
\pgfsetbuttcap%
\pgfsetroundjoin%
\definecolor{currentfill}{rgb}{0.000000,0.000000,0.000000}%
\pgfsetfillcolor{currentfill}%
\pgfsetlinewidth{0.803000pt}%
\definecolor{currentstroke}{rgb}{0.000000,0.000000,0.000000}%
\pgfsetstrokecolor{currentstroke}%
\pgfsetdash{}{0pt}%
\pgfsys@defobject{currentmarker}{\pgfqpoint{-0.048611in}{0.000000in}}{\pgfqpoint{-0.000000in}{0.000000in}}{%
\pgfpathmoveto{\pgfqpoint{-0.000000in}{0.000000in}}%
\pgfpathlineto{\pgfqpoint{-0.048611in}{0.000000in}}%
\pgfusepath{stroke,fill}%
}%
\begin{pgfscope}%
\pgfsys@transformshift{0.619136in}{0.955792in}%
\pgfsys@useobject{currentmarker}{}%
\end{pgfscope}%
\end{pgfscope}%
\begin{pgfscope}%
\definecolor{textcolor}{rgb}{0.000000,0.000000,0.000000}%
\pgfsetstrokecolor{textcolor}%
\pgfsetfillcolor{textcolor}%
\pgftext[x=0.344444in, y=0.907598in, left, base]{\color{textcolor}{\rmfamily\fontsize{10.000000}{12.000000}\selectfont\catcode`\^=\active\def^{\ifmmode\sp\else\^{}\fi}\catcode`\%=\active\def
\end{pgfscope}%
\begin{pgfscope}%
\pgfsetbuttcap%
\pgfsetroundjoin%
\definecolor{currentfill}{rgb}{0.000000,0.000000,0.000000}%
\pgfsetfillcolor{currentfill}%
\pgfsetlinewidth{0.803000pt}%
\definecolor{currentstroke}{rgb}{0.000000,0.000000,0.000000}%
\pgfsetstrokecolor{currentstroke}%
\pgfsetdash{}{0pt}%
\pgfsys@defobject{currentmarker}{\pgfqpoint{-0.048611in}{0.000000in}}{\pgfqpoint{-0.000000in}{0.000000in}}{%
\pgfpathmoveto{\pgfqpoint{-0.000000in}{0.000000in}}%
\pgfpathlineto{\pgfqpoint{-0.048611in}{0.000000in}}%
\pgfusepath{stroke,fill}%
}%
\begin{pgfscope}%
\pgfsys@transformshift{0.619136in}{1.275532in}%
\pgfsys@useobject{currentmarker}{}%
\end{pgfscope}%
\end{pgfscope}%
\begin{pgfscope}%
\definecolor{textcolor}{rgb}{0.000000,0.000000,0.000000}%
\pgfsetstrokecolor{textcolor}%
\pgfsetfillcolor{textcolor}%
\pgftext[x=0.344444in, y=1.227338in, left, base]{\color{textcolor}{\rmfamily\fontsize{10.000000}{12.000000}\selectfont\catcode`\^=\active\def^{\ifmmode\sp\else\^{}\fi}\catcode`\%=\active\def
\end{pgfscope}%
\begin{pgfscope}%
\pgfsetbuttcap%
\pgfsetroundjoin%
\definecolor{currentfill}{rgb}{0.000000,0.000000,0.000000}%
\pgfsetfillcolor{currentfill}%
\pgfsetlinewidth{0.803000pt}%
\definecolor{currentstroke}{rgb}{0.000000,0.000000,0.000000}%
\pgfsetstrokecolor{currentstroke}%
\pgfsetdash{}{0pt}%
\pgfsys@defobject{currentmarker}{\pgfqpoint{-0.048611in}{0.000000in}}{\pgfqpoint{-0.000000in}{0.000000in}}{%
\pgfpathmoveto{\pgfqpoint{-0.000000in}{0.000000in}}%
\pgfpathlineto{\pgfqpoint{-0.048611in}{0.000000in}}%
\pgfusepath{stroke,fill}%
}%
\begin{pgfscope}%
\pgfsys@transformshift{0.619136in}{1.595272in}%
\pgfsys@useobject{currentmarker}{}%
\end{pgfscope}%
\end{pgfscope}%
\begin{pgfscope}%
\definecolor{textcolor}{rgb}{0.000000,0.000000,0.000000}%
\pgfsetstrokecolor{textcolor}%
\pgfsetfillcolor{textcolor}%
\pgftext[x=0.452469in, y=1.547077in, left, base]{\color{textcolor}{\rmfamily\fontsize{10.000000}{12.000000}\selectfont\catcode`\^=\active\def^{\ifmmode\sp\else\^{}\fi}\catcode`\%=\active\def
\end{pgfscope}%
\begin{pgfscope}%
\pgfsetbuttcap%
\pgfsetroundjoin%
\definecolor{currentfill}{rgb}{0.000000,0.000000,0.000000}%
\pgfsetfillcolor{currentfill}%
\pgfsetlinewidth{0.803000pt}%
\definecolor{currentstroke}{rgb}{0.000000,0.000000,0.000000}%
\pgfsetstrokecolor{currentstroke}%
\pgfsetdash{}{0pt}%
\pgfsys@defobject{currentmarker}{\pgfqpoint{-0.048611in}{0.000000in}}{\pgfqpoint{-0.000000in}{0.000000in}}{%
\pgfpathmoveto{\pgfqpoint{-0.000000in}{0.000000in}}%
\pgfpathlineto{\pgfqpoint{-0.048611in}{0.000000in}}%
\pgfusepath{stroke,fill}%
}%
\begin{pgfscope}%
\pgfsys@transformshift{0.619136in}{1.915012in}%
\pgfsys@useobject{currentmarker}{}%
\end{pgfscope}%
\end{pgfscope}%
\begin{pgfscope}%
\definecolor{textcolor}{rgb}{0.000000,0.000000,0.000000}%
\pgfsetstrokecolor{textcolor}%
\pgfsetfillcolor{textcolor}%
\pgftext[x=0.452469in, y=1.866817in, left, base]{\color{textcolor}{\rmfamily\fontsize{10.000000}{12.000000}\selectfont\catcode`\^=\active\def^{\ifmmode\sp\else\^{}\fi}\catcode`\%=\active\def
\end{pgfscope}%
\begin{pgfscope}%
\definecolor{textcolor}{rgb}{0.000000,0.000000,0.000000}%
\pgfsetstrokecolor{textcolor}%
\pgfsetfillcolor{textcolor}%
\pgftext[x=0.288889in,y=1.355467in,,bottom,rotate=90.000000]{\color{textcolor}{\rmfamily\fontsize{10.000000}{12.000000}\selectfont\catcode`\^=\active\def^{\ifmmode\sp\else\^{}\fi}\catcode`\%=\active\def
\end{pgfscope}%
\begin{pgfscope}%
\pgfsetrectcap%
\pgfsetmiterjoin%
\pgfsetlinewidth{0.803000pt}%
\definecolor{currentstroke}{rgb}{0.000000,0.000000,0.000000}%
\pgfsetstrokecolor{currentstroke}%
\pgfsetdash{}{0pt}%
\pgfpathmoveto{\pgfqpoint{0.619136in}{0.556118in}}%
\pgfpathlineto{\pgfqpoint{0.619136in}{2.154816in}}%
\pgfusepath{stroke}%
\end{pgfscope}%
\begin{pgfscope}%
\pgfsetrectcap%
\pgfsetmiterjoin%
\pgfsetlinewidth{0.803000pt}%
\definecolor{currentstroke}{rgb}{0.000000,0.000000,0.000000}%
\pgfsetstrokecolor{currentstroke}%
\pgfsetdash{}{0pt}%
\pgfpathmoveto{\pgfqpoint{3.336924in}{0.556118in}}%
\pgfpathlineto{\pgfqpoint{3.336924in}{2.154816in}}%
\pgfusepath{stroke}%
\end{pgfscope}%
\begin{pgfscope}%
\pgfsetrectcap%
\pgfsetmiterjoin%
\pgfsetlinewidth{0.803000pt}%
\definecolor{currentstroke}{rgb}{0.000000,0.000000,0.000000}%
\pgfsetstrokecolor{currentstroke}%
\pgfsetdash{}{0pt}%
\pgfpathmoveto{\pgfqpoint{0.619136in}{0.556118in}}%
\pgfpathlineto{\pgfqpoint{3.336924in}{0.556118in}}%
\pgfusepath{stroke}%
\end{pgfscope}%
\begin{pgfscope}%
\pgfsetrectcap%
\pgfsetmiterjoin%
\pgfsetlinewidth{0.803000pt}%
\definecolor{currentstroke}{rgb}{0.000000,0.000000,0.000000}%
\pgfsetstrokecolor{currentstroke}%
\pgfsetdash{}{0pt}%
\pgfpathmoveto{\pgfqpoint{0.619136in}{2.154816in}}%
\pgfpathlineto{\pgfqpoint{3.336924in}{2.154816in}}%
\pgfusepath{stroke}%
\end{pgfscope}%
\end{pgfpicture}%
\makeatother%
\endgroup%

%% file: images/pulse.pgf
\providecommand{\mathdefault}[1]{#1}
\begingroup%
\makeatletter%
\begin{pgfpicture}%
\pgfpathrectangle{\pgfpointorigin}{\pgfqpoint{3.486924in}{2.789539in}}%
\pgfusepath{use as bounding box, clip}%
\begin{pgfscope}%
\pgfsetbuttcap%
\pgfsetmiterjoin%
\definecolor{currentfill}{rgb}{1.000000,1.000000,1.000000}%
\pgfsetfillcolor{currentfill}%
\pgfsetlinewidth{0.000000pt}%
\definecolor{currentstroke}{rgb}{1.000000,1.000000,1.000000}%
\pgfsetstrokecolor{currentstroke}%
\pgfsetdash{}{0pt}%
\pgfpathmoveto{\pgfqpoint{0.000000in}{0.000000in}}%
\pgfpathlineto{\pgfqpoint{3.486924in}{0.000000in}}%
\pgfpathlineto{\pgfqpoint{3.486924in}{2.789539in}}%
\pgfpathlineto{\pgfqpoint{0.000000in}{2.789539in}}%
\pgfpathlineto{\pgfqpoint{0.000000in}{0.000000in}}%
\pgfpathclose%
\pgfusepath{fill}%
\end{pgfscope}%
\begin{pgfscope}%
\pgfsetbuttcap%
\pgfsetmiterjoin%
\definecolor{currentfill}{rgb}{1.000000,1.000000,1.000000}%
\pgfsetfillcolor{currentfill}%
\pgfsetlinewidth{0.000000pt}%
\definecolor{currentstroke}{rgb}{0.000000,0.000000,0.000000}%
\pgfsetstrokecolor{currentstroke}%
\pgfsetstrokeopacity{0.000000}%
\pgfsetdash{}{0pt}%
\pgfpathmoveto{\pgfqpoint{0.682063in}{0.565000in}}%
\pgfpathlineto{\pgfqpoint{2.708408in}{0.565000in}}%
\pgfpathlineto{\pgfqpoint{2.708408in}{2.591345in}}%
\pgfpathlineto{\pgfqpoint{0.682063in}{2.591345in}}%
\pgfpathlineto{\pgfqpoint{0.682063in}{0.565000in}}%
\pgfpathclose%
\pgfusepath{fill}%
\end{pgfscope}%
\begin{pgfscope}%
\pgfpathrectangle{\pgfqpoint{0.682063in}{0.565000in}}{\pgfqpoint{2.026345in}{2.026345in}}%
\pgfusepath{clip}%
\pgfsys@transformcm{2.026345}{0.000000}{0.000000}{-2.026345}{0.682063in}{2.591345in}%
\pgftext[left,bottom]{\includegraphics[interpolate=false,width=1.000000in,height=1.000000in]{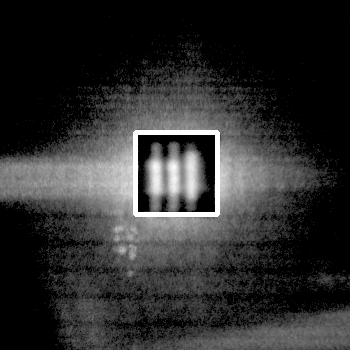}}%
\end{pgfscope}%
\begin{pgfscope}%
\pgfsetbuttcap%
\pgfsetroundjoin%
\definecolor{currentfill}{rgb}{0.000000,0.000000,0.000000}%
\pgfsetfillcolor{currentfill}%
\pgfsetlinewidth{0.803000pt}%
\definecolor{currentstroke}{rgb}{0.000000,0.000000,0.000000}%
\pgfsetstrokecolor{currentstroke}%
\pgfsetdash{}{0pt}%
\pgfsys@defobject{currentmarker}{\pgfqpoint{0.000000in}{-0.048611in}}{\pgfqpoint{0.000000in}{0.000000in}}{%
\pgfpathmoveto{\pgfqpoint{0.000000in}{0.000000in}}%
\pgfpathlineto{\pgfqpoint{0.000000in}{-0.048611in}}%
\pgfusepath{stroke,fill}%
}%
\begin{pgfscope}%
\pgfsys@transformshift{1.116280in}{0.565000in}%
\pgfsys@useobject{currentmarker}{}%
\end{pgfscope}%
\end{pgfscope}%
\begin{pgfscope}%
\definecolor{textcolor}{rgb}{0.000000,0.000000,0.000000}%
\pgfsetstrokecolor{textcolor}%
\pgfsetfillcolor{textcolor}%
\pgftext[x=1.116280in,y=0.467777in,,top]{\color{textcolor}{\rmfamily\fontsize{10.000000}{12.000000}\selectfont\catcode`\^=\active\def^{\ifmmode\sp\else\^{}\fi}\catcode`\%=\active\def
\end{pgfscope}%
\begin{pgfscope}%
\pgfsetbuttcap%
\pgfsetroundjoin%
\definecolor{currentfill}{rgb}{0.000000,0.000000,0.000000}%
\pgfsetfillcolor{currentfill}%
\pgfsetlinewidth{0.803000pt}%
\definecolor{currentstroke}{rgb}{0.000000,0.000000,0.000000}%
\pgfsetstrokecolor{currentstroke}%
\pgfsetdash{}{0pt}%
\pgfsys@defobject{currentmarker}{\pgfqpoint{0.000000in}{-0.048611in}}{\pgfqpoint{0.000000in}{0.000000in}}{%
\pgfpathmoveto{\pgfqpoint{0.000000in}{0.000000in}}%
\pgfpathlineto{\pgfqpoint{0.000000in}{-0.048611in}}%
\pgfusepath{stroke,fill}%
}%
\begin{pgfscope}%
\pgfsys@transformshift{1.695236in}{0.565000in}%
\pgfsys@useobject{currentmarker}{}%
\end{pgfscope}%
\end{pgfscope}%
\begin{pgfscope}%
\definecolor{textcolor}{rgb}{0.000000,0.000000,0.000000}%
\pgfsetstrokecolor{textcolor}%
\pgfsetfillcolor{textcolor}%
\pgftext[x=1.695236in,y=0.467777in,,top]{\color{textcolor}{\rmfamily\fontsize{10.000000}{12.000000}\selectfont\catcode`\^=\active\def^{\ifmmode\sp\else\^{}\fi}\catcode`\%=\active\def
\end{pgfscope}%
\begin{pgfscope}%
\pgfsetbuttcap%
\pgfsetroundjoin%
\definecolor{currentfill}{rgb}{0.000000,0.000000,0.000000}%
\pgfsetfillcolor{currentfill}%
\pgfsetlinewidth{0.803000pt}%
\definecolor{currentstroke}{rgb}{0.000000,0.000000,0.000000}%
\pgfsetstrokecolor{currentstroke}%
\pgfsetdash{}{0pt}%
\pgfsys@defobject{currentmarker}{\pgfqpoint{0.000000in}{-0.048611in}}{\pgfqpoint{0.000000in}{0.000000in}}{%
\pgfpathmoveto{\pgfqpoint{0.000000in}{0.000000in}}%
\pgfpathlineto{\pgfqpoint{0.000000in}{-0.048611in}}%
\pgfusepath{stroke,fill}%
}%
\begin{pgfscope}%
\pgfsys@transformshift{2.274191in}{0.565000in}%
\pgfsys@useobject{currentmarker}{}%
\end{pgfscope}%
\end{pgfscope}%
\begin{pgfscope}%
\definecolor{textcolor}{rgb}{0.000000,0.000000,0.000000}%
\pgfsetstrokecolor{textcolor}%
\pgfsetfillcolor{textcolor}%
\pgftext[x=2.274191in,y=0.467777in,,top]{\color{textcolor}{\rmfamily\fontsize{10.000000}{12.000000}\selectfont\catcode`\^=\active\def^{\ifmmode\sp\else\^{}\fi}\catcode`\%=\active\def
\end{pgfscope}%
\begin{pgfscope}%
\definecolor{textcolor}{rgb}{0.000000,0.000000,0.000000}%
\pgfsetstrokecolor{textcolor}%
\pgfsetfillcolor{textcolor}%
\pgftext[x=1.695236in,y=0.288889in,,top]{\color{textcolor}{\rmfamily\fontsize{10.000000}{12.000000}\selectfont\catcode`\^=\active\def^{\ifmmode\sp\else\^{}\fi}\catcode`\%=\active\def
\end{pgfscope}%
\begin{pgfscope}%
\pgfsetbuttcap%
\pgfsetroundjoin%
\definecolor{currentfill}{rgb}{0.000000,0.000000,0.000000}%
\pgfsetfillcolor{currentfill}%
\pgfsetlinewidth{0.803000pt}%
\definecolor{currentstroke}{rgb}{0.000000,0.000000,0.000000}%
\pgfsetstrokecolor{currentstroke}%
\pgfsetdash{}{0pt}%
\pgfsys@defobject{currentmarker}{\pgfqpoint{-0.048611in}{0.000000in}}{\pgfqpoint{-0.000000in}{0.000000in}}{%
\pgfpathmoveto{\pgfqpoint{-0.000000in}{0.000000in}}%
\pgfpathlineto{\pgfqpoint{-0.048611in}{0.000000in}}%
\pgfusepath{stroke,fill}%
}%
\begin{pgfscope}%
\pgfsys@transformshift{0.682063in}{0.999216in}%
\pgfsys@useobject{currentmarker}{}%
\end{pgfscope}%
\end{pgfscope}%
\begin{pgfscope}%
\definecolor{textcolor}{rgb}{0.000000,0.000000,0.000000}%
\pgfsetstrokecolor{textcolor}%
\pgfsetfillcolor{textcolor}%
\pgftext[x=0.469146in, y=0.951022in, left, base]{\color{textcolor}{\rmfamily\fontsize{10.000000}{12.000000}\selectfont\catcode`\^=\active\def^{\ifmmode\sp\else\^{}\fi}\catcode`\%=\active\def
\end{pgfscope}%
\begin{pgfscope}%
\pgfsetbuttcap%
\pgfsetroundjoin%
\definecolor{currentfill}{rgb}{0.000000,0.000000,0.000000}%
\pgfsetfillcolor{currentfill}%
\pgfsetlinewidth{0.803000pt}%
\definecolor{currentstroke}{rgb}{0.000000,0.000000,0.000000}%
\pgfsetstrokecolor{currentstroke}%
\pgfsetdash{}{0pt}%
\pgfsys@defobject{currentmarker}{\pgfqpoint{-0.048611in}{0.000000in}}{\pgfqpoint{-0.000000in}{0.000000in}}{%
\pgfpathmoveto{\pgfqpoint{-0.000000in}{0.000000in}}%
\pgfpathlineto{\pgfqpoint{-0.048611in}{0.000000in}}%
\pgfusepath{stroke,fill}%
}%
\begin{pgfscope}%
\pgfsys@transformshift{0.682063in}{1.578172in}%
\pgfsys@useobject{currentmarker}{}%
\end{pgfscope}%
\end{pgfscope}%
\begin{pgfscope}%
\definecolor{textcolor}{rgb}{0.000000,0.000000,0.000000}%
\pgfsetstrokecolor{textcolor}%
\pgfsetfillcolor{textcolor}%
\pgftext[x=0.515396in, y=1.529978in, left, base]{\color{textcolor}{\rmfamily\fontsize{10.000000}{12.000000}\selectfont\catcode`\^=\active\def^{\ifmmode\sp\else\^{}\fi}\catcode`\%=\active\def
\end{pgfscope}%
\begin{pgfscope}%
\pgfsetbuttcap%
\pgfsetroundjoin%
\definecolor{currentfill}{rgb}{0.000000,0.000000,0.000000}%
\pgfsetfillcolor{currentfill}%
\pgfsetlinewidth{0.803000pt}%
\definecolor{currentstroke}{rgb}{0.000000,0.000000,0.000000}%
\pgfsetstrokecolor{currentstroke}%
\pgfsetdash{}{0pt}%
\pgfsys@defobject{currentmarker}{\pgfqpoint{-0.048611in}{0.000000in}}{\pgfqpoint{-0.000000in}{0.000000in}}{%
\pgfpathmoveto{\pgfqpoint{-0.000000in}{0.000000in}}%
\pgfpathlineto{\pgfqpoint{-0.048611in}{0.000000in}}%
\pgfusepath{stroke,fill}%
}%
\begin{pgfscope}%
\pgfsys@transformshift{0.682063in}{2.157128in}%
\pgfsys@useobject{currentmarker}{}%
\end{pgfscope}%
\end{pgfscope}%
\begin{pgfscope}%
\definecolor{textcolor}{rgb}{0.000000,0.000000,0.000000}%
\pgfsetstrokecolor{textcolor}%
\pgfsetfillcolor{textcolor}%
\pgftext[x=0.515396in, y=2.108934in, left, base]{\color{textcolor}{\rmfamily\fontsize{10.000000}{12.000000}\selectfont\catcode`\^=\active\def^{\ifmmode\sp\else\^{}\fi}\catcode`\%=\active\def
\end{pgfscope}%
\begin{pgfscope}%
\definecolor{textcolor}{rgb}{0.000000,0.000000,0.000000}%
\pgfsetstrokecolor{textcolor}%
\pgfsetfillcolor{textcolor}%
\pgftext[x=0.413591in,y=1.578172in,,bottom,rotate=90.000000]{\color{textcolor}{\rmfamily\fontsize{10.000000}{12.000000}\selectfont\catcode`\^=\active\def^{\ifmmode\sp\else\^{}\fi}\catcode`\%=\active\def
\end{pgfscope}%
\begin{pgfscope}%
\pgfsetrectcap%
\pgfsetmiterjoin%
\pgfsetlinewidth{0.803000pt}%
\definecolor{currentstroke}{rgb}{0.000000,0.000000,0.000000}%
\pgfsetstrokecolor{currentstroke}%
\pgfsetdash{}{0pt}%
\pgfpathmoveto{\pgfqpoint{0.682063in}{0.565000in}}%
\pgfpathlineto{\pgfqpoint{0.682063in}{2.591345in}}%
\pgfusepath{stroke}%
\end{pgfscope}%
\begin{pgfscope}%
\pgfsetrectcap%
\pgfsetmiterjoin%
\pgfsetlinewidth{0.803000pt}%
\definecolor{currentstroke}{rgb}{0.000000,0.000000,0.000000}%
\pgfsetstrokecolor{currentstroke}%
\pgfsetdash{}{0pt}%
\pgfpathmoveto{\pgfqpoint{2.708408in}{0.565000in}}%
\pgfpathlineto{\pgfqpoint{2.708408in}{2.591345in}}%
\pgfusepath{stroke}%
\end{pgfscope}%
\begin{pgfscope}%
\pgfsetrectcap%
\pgfsetmiterjoin%
\pgfsetlinewidth{0.803000pt}%
\definecolor{currentstroke}{rgb}{0.000000,0.000000,0.000000}%
\pgfsetstrokecolor{currentstroke}%
\pgfsetdash{}{0pt}%
\pgfpathmoveto{\pgfqpoint{0.682063in}{0.565000in}}%
\pgfpathlineto{\pgfqpoint{2.708408in}{0.565000in}}%
\pgfusepath{stroke}%
\end{pgfscope}%
\begin{pgfscope}%
\pgfsetrectcap%
\pgfsetmiterjoin%
\pgfsetlinewidth{0.803000pt}%
\definecolor{currentstroke}{rgb}{0.000000,0.000000,0.000000}%
\pgfsetstrokecolor{currentstroke}%
\pgfsetdash{}{0pt}%
\pgfpathmoveto{\pgfqpoint{0.682063in}{2.591345in}}%
\pgfpathlineto{\pgfqpoint{2.708408in}{2.591345in}}%
\pgfusepath{stroke}%
\end{pgfscope}%
\begin{pgfscope}%
\pgfsetbuttcap%
\pgfsetmiterjoin%
\definecolor{currentfill}{rgb}{1.000000,1.000000,1.000000}%
\pgfsetfillcolor{currentfill}%
\pgfsetlinewidth{0.000000pt}%
\definecolor{currentstroke}{rgb}{0.000000,0.000000,0.000000}%
\pgfsetstrokecolor{currentstroke}%
\pgfsetstrokeopacity{0.000000}%
\pgfsetdash{}{0pt}%
\pgfpathmoveto{\pgfqpoint{2.843720in}{0.565000in}}%
\pgfpathlineto{\pgfqpoint{2.945038in}{0.565000in}}%
\pgfpathlineto{\pgfqpoint{2.945038in}{2.591345in}}%
\pgfpathlineto{\pgfqpoint{2.843720in}{2.591345in}}%
\pgfpathlineto{\pgfqpoint{2.843720in}{0.565000in}}%
\pgfpathclose%
\pgfusepath{fill}%
\end{pgfscope}%
\begin{pgfscope}%
\pgfsys@transformshift{2.840000in}{0.569539in}%
\pgftext[left,bottom]{\includegraphics[interpolate=true,width=0.110000in,height=2.030000in]{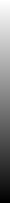}}%
\end{pgfscope}%
\begin{pgfscope}%
\pgfsetbuttcap%
\pgfsetroundjoin%
\definecolor{currentfill}{rgb}{0.000000,0.000000,0.000000}%
\pgfsetfillcolor{currentfill}%
\pgfsetlinewidth{0.803000pt}%
\definecolor{currentstroke}{rgb}{0.000000,0.000000,0.000000}%
\pgfsetstrokecolor{currentstroke}%
\pgfsetdash{}{0pt}%
\pgfsys@defobject{currentmarker}{\pgfqpoint{0.000000in}{0.000000in}}{\pgfqpoint{0.048611in}{0.000000in}}{%
\pgfpathmoveto{\pgfqpoint{0.000000in}{0.000000in}}%
\pgfpathlineto{\pgfqpoint{0.048611in}{0.000000in}}%
\pgfusepath{stroke,fill}%
}%
\begin{pgfscope}%
\pgfsys@transformshift{2.945038in}{0.565000in}%
\pgfsys@useobject{currentmarker}{}%
\end{pgfscope}%
\end{pgfscope}%
\begin{pgfscope}%
\definecolor{textcolor}{rgb}{0.000000,0.000000,0.000000}%
\pgfsetstrokecolor{textcolor}%
\pgfsetfillcolor{textcolor}%
\pgftext[x=3.042260in, y=0.516805in, left, base]{\color{textcolor}{\rmfamily\fontsize{10.000000}{12.000000}\selectfont\catcode`\^=\active\def^{\ifmmode\sp\else\^{}\fi}\catcode`\%=\active\def
\end{pgfscope}%
\begin{pgfscope}%
\pgfsetbuttcap%
\pgfsetroundjoin%
\definecolor{currentfill}{rgb}{0.000000,0.000000,0.000000}%
\pgfsetfillcolor{currentfill}%
\pgfsetlinewidth{0.803000pt}%
\definecolor{currentstroke}{rgb}{0.000000,0.000000,0.000000}%
\pgfsetstrokecolor{currentstroke}%
\pgfsetdash{}{0pt}%
\pgfsys@defobject{currentmarker}{\pgfqpoint{0.000000in}{0.000000in}}{\pgfqpoint{0.048611in}{0.000000in}}{%
\pgfpathmoveto{\pgfqpoint{0.000000in}{0.000000in}}%
\pgfpathlineto{\pgfqpoint{0.048611in}{0.000000in}}%
\pgfusepath{stroke,fill}%
}%
\begin{pgfscope}%
\pgfsys@transformshift{2.945038in}{1.240448in}%
\pgfsys@useobject{currentmarker}{}%
\end{pgfscope}%
\end{pgfscope}%
\begin{pgfscope}%
\definecolor{textcolor}{rgb}{0.000000,0.000000,0.000000}%
\pgfsetstrokecolor{textcolor}%
\pgfsetfillcolor{textcolor}%
\pgftext[x=3.042260in, y=1.192254in, left, base]{\color{textcolor}{\rmfamily\fontsize{10.000000}{12.000000}\selectfont\catcode`\^=\active\def^{\ifmmode\sp\else\^{}\fi}\catcode`\%=\active\def
\end{pgfscope}%
\begin{pgfscope}%
\pgfsetbuttcap%
\pgfsetroundjoin%
\definecolor{currentfill}{rgb}{0.000000,0.000000,0.000000}%
\pgfsetfillcolor{currentfill}%
\pgfsetlinewidth{0.803000pt}%
\definecolor{currentstroke}{rgb}{0.000000,0.000000,0.000000}%
\pgfsetstrokecolor{currentstroke}%
\pgfsetdash{}{0pt}%
\pgfsys@defobject{currentmarker}{\pgfqpoint{0.000000in}{0.000000in}}{\pgfqpoint{0.048611in}{0.000000in}}{%
\pgfpathmoveto{\pgfqpoint{0.000000in}{0.000000in}}%
\pgfpathlineto{\pgfqpoint{0.048611in}{0.000000in}}%
\pgfusepath{stroke,fill}%
}%
\begin{pgfscope}%
\pgfsys@transformshift{2.945038in}{1.915896in}%
\pgfsys@useobject{currentmarker}{}%
\end{pgfscope}%
\end{pgfscope}%
\begin{pgfscope}%
\definecolor{textcolor}{rgb}{0.000000,0.000000,0.000000}%
\pgfsetstrokecolor{textcolor}%
\pgfsetfillcolor{textcolor}%
\pgftext[x=3.042260in, y=1.867702in, left, base]{\color{textcolor}{\rmfamily\fontsize{10.000000}{12.000000}\selectfont\catcode`\^=\active\def^{\ifmmode\sp\else\^{}\fi}\catcode`\%=\active\def
\end{pgfscope}%
\begin{pgfscope}%
\pgfsetbuttcap%
\pgfsetroundjoin%
\definecolor{currentfill}{rgb}{0.000000,0.000000,0.000000}%
\pgfsetfillcolor{currentfill}%
\pgfsetlinewidth{0.803000pt}%
\definecolor{currentstroke}{rgb}{0.000000,0.000000,0.000000}%
\pgfsetstrokecolor{currentstroke}%
\pgfsetdash{}{0pt}%
\pgfsys@defobject{currentmarker}{\pgfqpoint{0.000000in}{0.000000in}}{\pgfqpoint{0.048611in}{0.000000in}}{%
\pgfpathmoveto{\pgfqpoint{0.000000in}{0.000000in}}%
\pgfpathlineto{\pgfqpoint{0.048611in}{0.000000in}}%
\pgfusepath{stroke,fill}%
}%
\begin{pgfscope}%
\pgfsys@transformshift{2.945038in}{2.591345in}%
\pgfsys@useobject{currentmarker}{}%
\end{pgfscope}%
\end{pgfscope}%
\begin{pgfscope}%
\definecolor{textcolor}{rgb}{0.000000,0.000000,0.000000}%
\pgfsetstrokecolor{textcolor}%
\pgfsetfillcolor{textcolor}%
\pgftext[x=3.042260in, y=2.543150in, left, base]{\color{textcolor}{\rmfamily\fontsize{10.000000}{12.000000}\selectfont\catcode`\^=\active\def^{\ifmmode\sp\else\^{}\fi}\catcode`\%=\active\def
\end{pgfscope}%
\begin{pgfscope}%
\pgfsetbuttcap%
\pgfsetroundjoin%
\definecolor{currentfill}{rgb}{0.000000,0.000000,0.000000}%
\pgfsetfillcolor{currentfill}%
\pgfsetlinewidth{0.602250pt}%
\definecolor{currentstroke}{rgb}{0.000000,0.000000,0.000000}%
\pgfsetstrokecolor{currentstroke}%
\pgfsetdash{}{0pt}%
\pgfsys@defobject{currentmarker}{\pgfqpoint{0.000000in}{0.000000in}}{\pgfqpoint{0.027778in}{0.000000in}}{%
\pgfpathmoveto{\pgfqpoint{0.000000in}{0.000000in}}%
\pgfpathlineto{\pgfqpoint{0.027778in}{0.000000in}}%
\pgfusepath{stroke,fill}%
}%
\begin{pgfscope}%
\pgfsys@transformshift{2.945038in}{0.768330in}%
\pgfsys@useobject{currentmarker}{}%
\end{pgfscope}%
\end{pgfscope}%
\begin{pgfscope}%
\pgfsetbuttcap%
\pgfsetroundjoin%
\definecolor{currentfill}{rgb}{0.000000,0.000000,0.000000}%
\pgfsetfillcolor{currentfill}%
\pgfsetlinewidth{0.602250pt}%
\definecolor{currentstroke}{rgb}{0.000000,0.000000,0.000000}%
\pgfsetstrokecolor{currentstroke}%
\pgfsetdash{}{0pt}%
\pgfsys@defobject{currentmarker}{\pgfqpoint{0.000000in}{0.000000in}}{\pgfqpoint{0.027778in}{0.000000in}}{%
\pgfpathmoveto{\pgfqpoint{0.000000in}{0.000000in}}%
\pgfpathlineto{\pgfqpoint{0.027778in}{0.000000in}}%
\pgfusepath{stroke,fill}%
}%
\begin{pgfscope}%
\pgfsys@transformshift{2.945038in}{0.887270in}%
\pgfsys@useobject{currentmarker}{}%
\end{pgfscope}%
\end{pgfscope}%
\begin{pgfscope}%
\pgfsetbuttcap%
\pgfsetroundjoin%
\definecolor{currentfill}{rgb}{0.000000,0.000000,0.000000}%
\pgfsetfillcolor{currentfill}%
\pgfsetlinewidth{0.602250pt}%
\definecolor{currentstroke}{rgb}{0.000000,0.000000,0.000000}%
\pgfsetstrokecolor{currentstroke}%
\pgfsetdash{}{0pt}%
\pgfsys@defobject{currentmarker}{\pgfqpoint{0.000000in}{0.000000in}}{\pgfqpoint{0.027778in}{0.000000in}}{%
\pgfpathmoveto{\pgfqpoint{0.000000in}{0.000000in}}%
\pgfpathlineto{\pgfqpoint{0.027778in}{0.000000in}}%
\pgfusepath{stroke,fill}%
}%
\begin{pgfscope}%
\pgfsys@transformshift{2.945038in}{0.971660in}%
\pgfsys@useobject{currentmarker}{}%
\end{pgfscope}%
\end{pgfscope}%
\begin{pgfscope}%
\pgfsetbuttcap%
\pgfsetroundjoin%
\definecolor{currentfill}{rgb}{0.000000,0.000000,0.000000}%
\pgfsetfillcolor{currentfill}%
\pgfsetlinewidth{0.602250pt}%
\definecolor{currentstroke}{rgb}{0.000000,0.000000,0.000000}%
\pgfsetstrokecolor{currentstroke}%
\pgfsetdash{}{0pt}%
\pgfsys@defobject{currentmarker}{\pgfqpoint{0.000000in}{0.000000in}}{\pgfqpoint{0.027778in}{0.000000in}}{%
\pgfpathmoveto{\pgfqpoint{0.000000in}{0.000000in}}%
\pgfpathlineto{\pgfqpoint{0.027778in}{0.000000in}}%
\pgfusepath{stroke,fill}%
}%
\begin{pgfscope}%
\pgfsys@transformshift{2.945038in}{1.037118in}%
\pgfsys@useobject{currentmarker}{}%
\end{pgfscope}%
\end{pgfscope}%
\begin{pgfscope}%
\pgfsetbuttcap%
\pgfsetroundjoin%
\definecolor{currentfill}{rgb}{0.000000,0.000000,0.000000}%
\pgfsetfillcolor{currentfill}%
\pgfsetlinewidth{0.602250pt}%
\definecolor{currentstroke}{rgb}{0.000000,0.000000,0.000000}%
\pgfsetstrokecolor{currentstroke}%
\pgfsetdash{}{0pt}%
\pgfsys@defobject{currentmarker}{\pgfqpoint{0.000000in}{0.000000in}}{\pgfqpoint{0.027778in}{0.000000in}}{%
\pgfpathmoveto{\pgfqpoint{0.000000in}{0.000000in}}%
\pgfpathlineto{\pgfqpoint{0.027778in}{0.000000in}}%
\pgfusepath{stroke,fill}%
}%
\begin{pgfscope}%
\pgfsys@transformshift{2.945038in}{1.090601in}%
\pgfsys@useobject{currentmarker}{}%
\end{pgfscope}%
\end{pgfscope}%
\begin{pgfscope}%
\pgfsetbuttcap%
\pgfsetroundjoin%
\definecolor{currentfill}{rgb}{0.000000,0.000000,0.000000}%
\pgfsetfillcolor{currentfill}%
\pgfsetlinewidth{0.602250pt}%
\definecolor{currentstroke}{rgb}{0.000000,0.000000,0.000000}%
\pgfsetstrokecolor{currentstroke}%
\pgfsetdash{}{0pt}%
\pgfsys@defobject{currentmarker}{\pgfqpoint{0.000000in}{0.000000in}}{\pgfqpoint{0.027778in}{0.000000in}}{%
\pgfpathmoveto{\pgfqpoint{0.000000in}{0.000000in}}%
\pgfpathlineto{\pgfqpoint{0.027778in}{0.000000in}}%
\pgfusepath{stroke,fill}%
}%
\begin{pgfscope}%
\pgfsys@transformshift{2.945038in}{1.135820in}%
\pgfsys@useobject{currentmarker}{}%
\end{pgfscope}%
\end{pgfscope}%
\begin{pgfscope}%
\pgfsetbuttcap%
\pgfsetroundjoin%
\definecolor{currentfill}{rgb}{0.000000,0.000000,0.000000}%
\pgfsetfillcolor{currentfill}%
\pgfsetlinewidth{0.602250pt}%
\definecolor{currentstroke}{rgb}{0.000000,0.000000,0.000000}%
\pgfsetstrokecolor{currentstroke}%
\pgfsetdash{}{0pt}%
\pgfsys@defobject{currentmarker}{\pgfqpoint{0.000000in}{0.000000in}}{\pgfqpoint{0.027778in}{0.000000in}}{%
\pgfpathmoveto{\pgfqpoint{0.000000in}{0.000000in}}%
\pgfpathlineto{\pgfqpoint{0.027778in}{0.000000in}}%
\pgfusepath{stroke,fill}%
}%
\begin{pgfscope}%
\pgfsys@transformshift{2.945038in}{1.174990in}%
\pgfsys@useobject{currentmarker}{}%
\end{pgfscope}%
\end{pgfscope}%
\begin{pgfscope}%
\pgfsetbuttcap%
\pgfsetroundjoin%
\definecolor{currentfill}{rgb}{0.000000,0.000000,0.000000}%
\pgfsetfillcolor{currentfill}%
\pgfsetlinewidth{0.602250pt}%
\definecolor{currentstroke}{rgb}{0.000000,0.000000,0.000000}%
\pgfsetstrokecolor{currentstroke}%
\pgfsetdash{}{0pt}%
\pgfsys@defobject{currentmarker}{\pgfqpoint{0.000000in}{0.000000in}}{\pgfqpoint{0.027778in}{0.000000in}}{%
\pgfpathmoveto{\pgfqpoint{0.000000in}{0.000000in}}%
\pgfpathlineto{\pgfqpoint{0.027778in}{0.000000in}}%
\pgfusepath{stroke,fill}%
}%
\begin{pgfscope}%
\pgfsys@transformshift{2.945038in}{1.209541in}%
\pgfsys@useobject{currentmarker}{}%
\end{pgfscope}%
\end{pgfscope}%
\begin{pgfscope}%
\pgfsetbuttcap%
\pgfsetroundjoin%
\definecolor{currentfill}{rgb}{0.000000,0.000000,0.000000}%
\pgfsetfillcolor{currentfill}%
\pgfsetlinewidth{0.602250pt}%
\definecolor{currentstroke}{rgb}{0.000000,0.000000,0.000000}%
\pgfsetstrokecolor{currentstroke}%
\pgfsetdash{}{0pt}%
\pgfsys@defobject{currentmarker}{\pgfqpoint{0.000000in}{0.000000in}}{\pgfqpoint{0.027778in}{0.000000in}}{%
\pgfpathmoveto{\pgfqpoint{0.000000in}{0.000000in}}%
\pgfpathlineto{\pgfqpoint{0.027778in}{0.000000in}}%
\pgfusepath{stroke,fill}%
}%
\begin{pgfscope}%
\pgfsys@transformshift{2.945038in}{1.443778in}%
\pgfsys@useobject{currentmarker}{}%
\end{pgfscope}%
\end{pgfscope}%
\begin{pgfscope}%
\pgfsetbuttcap%
\pgfsetroundjoin%
\definecolor{currentfill}{rgb}{0.000000,0.000000,0.000000}%
\pgfsetfillcolor{currentfill}%
\pgfsetlinewidth{0.602250pt}%
\definecolor{currentstroke}{rgb}{0.000000,0.000000,0.000000}%
\pgfsetstrokecolor{currentstroke}%
\pgfsetdash{}{0pt}%
\pgfsys@defobject{currentmarker}{\pgfqpoint{0.000000in}{0.000000in}}{\pgfqpoint{0.027778in}{0.000000in}}{%
\pgfpathmoveto{\pgfqpoint{0.000000in}{0.000000in}}%
\pgfpathlineto{\pgfqpoint{0.027778in}{0.000000in}}%
\pgfusepath{stroke,fill}%
}%
\begin{pgfscope}%
\pgfsys@transformshift{2.945038in}{1.562719in}%
\pgfsys@useobject{currentmarker}{}%
\end{pgfscope}%
\end{pgfscope}%
\begin{pgfscope}%
\pgfsetbuttcap%
\pgfsetroundjoin%
\definecolor{currentfill}{rgb}{0.000000,0.000000,0.000000}%
\pgfsetfillcolor{currentfill}%
\pgfsetlinewidth{0.602250pt}%
\definecolor{currentstroke}{rgb}{0.000000,0.000000,0.000000}%
\pgfsetstrokecolor{currentstroke}%
\pgfsetdash{}{0pt}%
\pgfsys@defobject{currentmarker}{\pgfqpoint{0.000000in}{0.000000in}}{\pgfqpoint{0.027778in}{0.000000in}}{%
\pgfpathmoveto{\pgfqpoint{0.000000in}{0.000000in}}%
\pgfpathlineto{\pgfqpoint{0.027778in}{0.000000in}}%
\pgfusepath{stroke,fill}%
}%
\begin{pgfscope}%
\pgfsys@transformshift{2.945038in}{1.647108in}%
\pgfsys@useobject{currentmarker}{}%
\end{pgfscope}%
\end{pgfscope}%
\begin{pgfscope}%
\pgfsetbuttcap%
\pgfsetroundjoin%
\definecolor{currentfill}{rgb}{0.000000,0.000000,0.000000}%
\pgfsetfillcolor{currentfill}%
\pgfsetlinewidth{0.602250pt}%
\definecolor{currentstroke}{rgb}{0.000000,0.000000,0.000000}%
\pgfsetstrokecolor{currentstroke}%
\pgfsetdash{}{0pt}%
\pgfsys@defobject{currentmarker}{\pgfqpoint{0.000000in}{0.000000in}}{\pgfqpoint{0.027778in}{0.000000in}}{%
\pgfpathmoveto{\pgfqpoint{0.000000in}{0.000000in}}%
\pgfpathlineto{\pgfqpoint{0.027778in}{0.000000in}}%
\pgfusepath{stroke,fill}%
}%
\begin{pgfscope}%
\pgfsys@transformshift{2.945038in}{1.712566in}%
\pgfsys@useobject{currentmarker}{}%
\end{pgfscope}%
\end{pgfscope}%
\begin{pgfscope}%
\pgfsetbuttcap%
\pgfsetroundjoin%
\definecolor{currentfill}{rgb}{0.000000,0.000000,0.000000}%
\pgfsetfillcolor{currentfill}%
\pgfsetlinewidth{0.602250pt}%
\definecolor{currentstroke}{rgb}{0.000000,0.000000,0.000000}%
\pgfsetstrokecolor{currentstroke}%
\pgfsetdash{}{0pt}%
\pgfsys@defobject{currentmarker}{\pgfqpoint{0.000000in}{0.000000in}}{\pgfqpoint{0.027778in}{0.000000in}}{%
\pgfpathmoveto{\pgfqpoint{0.000000in}{0.000000in}}%
\pgfpathlineto{\pgfqpoint{0.027778in}{0.000000in}}%
\pgfusepath{stroke,fill}%
}%
\begin{pgfscope}%
\pgfsys@transformshift{2.945038in}{1.766049in}%
\pgfsys@useobject{currentmarker}{}%
\end{pgfscope}%
\end{pgfscope}%
\begin{pgfscope}%
\pgfsetbuttcap%
\pgfsetroundjoin%
\definecolor{currentfill}{rgb}{0.000000,0.000000,0.000000}%
\pgfsetfillcolor{currentfill}%
\pgfsetlinewidth{0.602250pt}%
\definecolor{currentstroke}{rgb}{0.000000,0.000000,0.000000}%
\pgfsetstrokecolor{currentstroke}%
\pgfsetdash{}{0pt}%
\pgfsys@defobject{currentmarker}{\pgfqpoint{0.000000in}{0.000000in}}{\pgfqpoint{0.027778in}{0.000000in}}{%
\pgfpathmoveto{\pgfqpoint{0.000000in}{0.000000in}}%
\pgfpathlineto{\pgfqpoint{0.027778in}{0.000000in}}%
\pgfusepath{stroke,fill}%
}%
\begin{pgfscope}%
\pgfsys@transformshift{2.945038in}{1.811268in}%
\pgfsys@useobject{currentmarker}{}%
\end{pgfscope}%
\end{pgfscope}%
\begin{pgfscope}%
\pgfsetbuttcap%
\pgfsetroundjoin%
\definecolor{currentfill}{rgb}{0.000000,0.000000,0.000000}%
\pgfsetfillcolor{currentfill}%
\pgfsetlinewidth{0.602250pt}%
\definecolor{currentstroke}{rgb}{0.000000,0.000000,0.000000}%
\pgfsetstrokecolor{currentstroke}%
\pgfsetdash{}{0pt}%
\pgfsys@defobject{currentmarker}{\pgfqpoint{0.000000in}{0.000000in}}{\pgfqpoint{0.027778in}{0.000000in}}{%
\pgfpathmoveto{\pgfqpoint{0.000000in}{0.000000in}}%
\pgfpathlineto{\pgfqpoint{0.027778in}{0.000000in}}%
\pgfusepath{stroke,fill}%
}%
\begin{pgfscope}%
\pgfsys@transformshift{2.945038in}{1.850439in}%
\pgfsys@useobject{currentmarker}{}%
\end{pgfscope}%
\end{pgfscope}%
\begin{pgfscope}%
\pgfsetbuttcap%
\pgfsetroundjoin%
\definecolor{currentfill}{rgb}{0.000000,0.000000,0.000000}%
\pgfsetfillcolor{currentfill}%
\pgfsetlinewidth{0.602250pt}%
\definecolor{currentstroke}{rgb}{0.000000,0.000000,0.000000}%
\pgfsetstrokecolor{currentstroke}%
\pgfsetdash{}{0pt}%
\pgfsys@defobject{currentmarker}{\pgfqpoint{0.000000in}{0.000000in}}{\pgfqpoint{0.027778in}{0.000000in}}{%
\pgfpathmoveto{\pgfqpoint{0.000000in}{0.000000in}}%
\pgfpathlineto{\pgfqpoint{0.027778in}{0.000000in}}%
\pgfusepath{stroke,fill}%
}%
\begin{pgfscope}%
\pgfsys@transformshift{2.945038in}{1.884990in}%
\pgfsys@useobject{currentmarker}{}%
\end{pgfscope}%
\end{pgfscope}%
\begin{pgfscope}%
\pgfsetbuttcap%
\pgfsetroundjoin%
\definecolor{currentfill}{rgb}{0.000000,0.000000,0.000000}%
\pgfsetfillcolor{currentfill}%
\pgfsetlinewidth{0.602250pt}%
\definecolor{currentstroke}{rgb}{0.000000,0.000000,0.000000}%
\pgfsetstrokecolor{currentstroke}%
\pgfsetdash{}{0pt}%
\pgfsys@defobject{currentmarker}{\pgfqpoint{0.000000in}{0.000000in}}{\pgfqpoint{0.027778in}{0.000000in}}{%
\pgfpathmoveto{\pgfqpoint{0.000000in}{0.000000in}}%
\pgfpathlineto{\pgfqpoint{0.027778in}{0.000000in}}%
\pgfusepath{stroke,fill}%
}%
\begin{pgfscope}%
\pgfsys@transformshift{2.945038in}{2.119227in}%
\pgfsys@useobject{currentmarker}{}%
\end{pgfscope}%
\end{pgfscope}%
\begin{pgfscope}%
\pgfsetbuttcap%
\pgfsetroundjoin%
\definecolor{currentfill}{rgb}{0.000000,0.000000,0.000000}%
\pgfsetfillcolor{currentfill}%
\pgfsetlinewidth{0.602250pt}%
\definecolor{currentstroke}{rgb}{0.000000,0.000000,0.000000}%
\pgfsetstrokecolor{currentstroke}%
\pgfsetdash{}{0pt}%
\pgfsys@defobject{currentmarker}{\pgfqpoint{0.000000in}{0.000000in}}{\pgfqpoint{0.027778in}{0.000000in}}{%
\pgfpathmoveto{\pgfqpoint{0.000000in}{0.000000in}}%
\pgfpathlineto{\pgfqpoint{0.027778in}{0.000000in}}%
\pgfusepath{stroke,fill}%
}%
\begin{pgfscope}%
\pgfsys@transformshift{2.945038in}{2.238167in}%
\pgfsys@useobject{currentmarker}{}%
\end{pgfscope}%
\end{pgfscope}%
\begin{pgfscope}%
\pgfsetbuttcap%
\pgfsetroundjoin%
\definecolor{currentfill}{rgb}{0.000000,0.000000,0.000000}%
\pgfsetfillcolor{currentfill}%
\pgfsetlinewidth{0.602250pt}%
\definecolor{currentstroke}{rgb}{0.000000,0.000000,0.000000}%
\pgfsetstrokecolor{currentstroke}%
\pgfsetdash{}{0pt}%
\pgfsys@defobject{currentmarker}{\pgfqpoint{0.000000in}{0.000000in}}{\pgfqpoint{0.027778in}{0.000000in}}{%
\pgfpathmoveto{\pgfqpoint{0.000000in}{0.000000in}}%
\pgfpathlineto{\pgfqpoint{0.027778in}{0.000000in}}%
\pgfusepath{stroke,fill}%
}%
\begin{pgfscope}%
\pgfsys@transformshift{2.945038in}{2.322557in}%
\pgfsys@useobject{currentmarker}{}%
\end{pgfscope}%
\end{pgfscope}%
\begin{pgfscope}%
\pgfsetbuttcap%
\pgfsetroundjoin%
\definecolor{currentfill}{rgb}{0.000000,0.000000,0.000000}%
\pgfsetfillcolor{currentfill}%
\pgfsetlinewidth{0.602250pt}%
\definecolor{currentstroke}{rgb}{0.000000,0.000000,0.000000}%
\pgfsetstrokecolor{currentstroke}%
\pgfsetdash{}{0pt}%
\pgfsys@defobject{currentmarker}{\pgfqpoint{0.000000in}{0.000000in}}{\pgfqpoint{0.027778in}{0.000000in}}{%
\pgfpathmoveto{\pgfqpoint{0.000000in}{0.000000in}}%
\pgfpathlineto{\pgfqpoint{0.027778in}{0.000000in}}%
\pgfusepath{stroke,fill}%
}%
\begin{pgfscope}%
\pgfsys@transformshift{2.945038in}{2.388015in}%
\pgfsys@useobject{currentmarker}{}%
\end{pgfscope}%
\end{pgfscope}%
\begin{pgfscope}%
\pgfsetbuttcap%
\pgfsetroundjoin%
\definecolor{currentfill}{rgb}{0.000000,0.000000,0.000000}%
\pgfsetfillcolor{currentfill}%
\pgfsetlinewidth{0.602250pt}%
\definecolor{currentstroke}{rgb}{0.000000,0.000000,0.000000}%
\pgfsetstrokecolor{currentstroke}%
\pgfsetdash{}{0pt}%
\pgfsys@defobject{currentmarker}{\pgfqpoint{0.000000in}{0.000000in}}{\pgfqpoint{0.027778in}{0.000000in}}{%
\pgfpathmoveto{\pgfqpoint{0.000000in}{0.000000in}}%
\pgfpathlineto{\pgfqpoint{0.027778in}{0.000000in}}%
\pgfusepath{stroke,fill}%
}%
\begin{pgfscope}%
\pgfsys@transformshift{2.945038in}{2.441497in}%
\pgfsys@useobject{currentmarker}{}%
\end{pgfscope}%
\end{pgfscope}%
\begin{pgfscope}%
\pgfsetbuttcap%
\pgfsetroundjoin%
\definecolor{currentfill}{rgb}{0.000000,0.000000,0.000000}%
\pgfsetfillcolor{currentfill}%
\pgfsetlinewidth{0.602250pt}%
\definecolor{currentstroke}{rgb}{0.000000,0.000000,0.000000}%
\pgfsetstrokecolor{currentstroke}%
\pgfsetdash{}{0pt}%
\pgfsys@defobject{currentmarker}{\pgfqpoint{0.000000in}{0.000000in}}{\pgfqpoint{0.027778in}{0.000000in}}{%
\pgfpathmoveto{\pgfqpoint{0.000000in}{0.000000in}}%
\pgfpathlineto{\pgfqpoint{0.027778in}{0.000000in}}%
\pgfusepath{stroke,fill}%
}%
\begin{pgfscope}%
\pgfsys@transformshift{2.945038in}{2.486717in}%
\pgfsys@useobject{currentmarker}{}%
\end{pgfscope}%
\end{pgfscope}%
\begin{pgfscope}%
\pgfsetbuttcap%
\pgfsetroundjoin%
\definecolor{currentfill}{rgb}{0.000000,0.000000,0.000000}%
\pgfsetfillcolor{currentfill}%
\pgfsetlinewidth{0.602250pt}%
\definecolor{currentstroke}{rgb}{0.000000,0.000000,0.000000}%
\pgfsetstrokecolor{currentstroke}%
\pgfsetdash{}{0pt}%
\pgfsys@defobject{currentmarker}{\pgfqpoint{0.000000in}{0.000000in}}{\pgfqpoint{0.027778in}{0.000000in}}{%
\pgfpathmoveto{\pgfqpoint{0.000000in}{0.000000in}}%
\pgfpathlineto{\pgfqpoint{0.027778in}{0.000000in}}%
\pgfusepath{stroke,fill}%
}%
\begin{pgfscope}%
\pgfsys@transformshift{2.945038in}{2.525887in}%
\pgfsys@useobject{currentmarker}{}%
\end{pgfscope}%
\end{pgfscope}%
\begin{pgfscope}%
\pgfsetbuttcap%
\pgfsetroundjoin%
\definecolor{currentfill}{rgb}{0.000000,0.000000,0.000000}%
\pgfsetfillcolor{currentfill}%
\pgfsetlinewidth{0.602250pt}%
\definecolor{currentstroke}{rgb}{0.000000,0.000000,0.000000}%
\pgfsetstrokecolor{currentstroke}%
\pgfsetdash{}{0pt}%
\pgfsys@defobject{currentmarker}{\pgfqpoint{0.000000in}{0.000000in}}{\pgfqpoint{0.027778in}{0.000000in}}{%
\pgfpathmoveto{\pgfqpoint{0.000000in}{0.000000in}}%
\pgfpathlineto{\pgfqpoint{0.027778in}{0.000000in}}%
\pgfusepath{stroke,fill}%
}%
\begin{pgfscope}%
\pgfsys@transformshift{2.945038in}{2.560438in}%
\pgfsys@useobject{currentmarker}{}%
\end{pgfscope}%
\end{pgfscope}%
\begin{pgfscope}%
\pgfsetrectcap%
\pgfsetmiterjoin%
\pgfsetlinewidth{0.803000pt}%
\definecolor{currentstroke}{rgb}{0.000000,0.000000,0.000000}%
\pgfsetstrokecolor{currentstroke}%
\pgfsetdash{}{0pt}%
\pgfpathmoveto{\pgfqpoint{2.843720in}{0.565000in}}%
\pgfpathlineto{\pgfqpoint{2.894379in}{0.565000in}}%
\pgfpathlineto{\pgfqpoint{2.945038in}{0.565000in}}%
\pgfpathlineto{\pgfqpoint{2.945038in}{2.591345in}}%
\pgfpathlineto{\pgfqpoint{2.894379in}{2.591345in}}%
\pgfpathlineto{\pgfqpoint{2.843720in}{2.591345in}}%
\pgfpathlineto{\pgfqpoint{2.843720in}{0.565000in}}%
\pgfpathclose%
\pgfusepath{stroke}%
\end{pgfscope}%
\end{pgfpicture}%
\makeatother%
\endgroup%

%% file: images/sensitivity.pgf
\providecommand{\mathdefault}[1]{#1}
\begingroup%
\makeatletter%
\begin{pgfpicture}%
\pgfpathrectangle{\pgfpointorigin}{\pgfqpoint{3.486924in}{2.789539in}}%
\pgfusepath{use as bounding box, clip}%
\begin{pgfscope}%
\pgfsetbuttcap%
\pgfsetmiterjoin%
\definecolor{currentfill}{rgb}{1.000000,1.000000,1.000000}%
\pgfsetfillcolor{currentfill}%
\pgfsetlinewidth{0.000000pt}%
\definecolor{currentstroke}{rgb}{1.000000,1.000000,1.000000}%
\pgfsetstrokecolor{currentstroke}%
\pgfsetdash{}{0pt}%
\pgfpathmoveto{\pgfqpoint{0.000000in}{0.000000in}}%
\pgfpathlineto{\pgfqpoint{3.486924in}{0.000000in}}%
\pgfpathlineto{\pgfqpoint{3.486924in}{2.789539in}}%
\pgfpathlineto{\pgfqpoint{0.000000in}{2.789539in}}%
\pgfpathlineto{\pgfqpoint{0.000000in}{0.000000in}}%
\pgfpathclose%
\pgfusepath{fill}%
\end{pgfscope}%
\begin{pgfscope}%
\pgfsetbuttcap%
\pgfsetmiterjoin%
\definecolor{currentfill}{rgb}{1.000000,1.000000,1.000000}%
\pgfsetfillcolor{currentfill}%
\pgfsetlinewidth{0.000000pt}%
\definecolor{currentstroke}{rgb}{0.000000,0.000000,0.000000}%
\pgfsetstrokecolor{currentstroke}%
\pgfsetstrokeopacity{0.000000}%
\pgfsetdash{}{0pt}%
\pgfpathmoveto{\pgfqpoint{0.785837in}{0.565000in}}%
\pgfpathlineto{\pgfqpoint{2.812182in}{0.565000in}}%
\pgfpathlineto{\pgfqpoint{2.812182in}{2.591345in}}%
\pgfpathlineto{\pgfqpoint{0.785837in}{2.591345in}}%
\pgfpathlineto{\pgfqpoint{0.785837in}{0.565000in}}%
\pgfpathclose%
\pgfusepath{fill}%
\end{pgfscope}%
\begin{pgfscope}%
\pgfpathrectangle{\pgfqpoint{0.785837in}{0.565000in}}{\pgfqpoint{2.026345in}{2.026345in}}%
\pgfusepath{clip}%
\pgfsys@transformcm{2.026345}{0.000000}{0.000000}{-2.026345}{0.785837in}{2.591345in}%
\pgftext[left,bottom]{\includegraphics[interpolate=false,width=1.000000in,height=1.000000in]{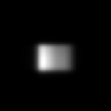}}%
\end{pgfscope}%
\begin{pgfscope}%
\pgfsetbuttcap%
\pgfsetroundjoin%
\definecolor{currentfill}{rgb}{0.000000,0.000000,0.000000}%
\pgfsetfillcolor{currentfill}%
\pgfsetlinewidth{0.803000pt}%
\definecolor{currentstroke}{rgb}{0.000000,0.000000,0.000000}%
\pgfsetstrokecolor{currentstroke}%
\pgfsetdash{}{0pt}%
\pgfsys@defobject{currentmarker}{\pgfqpoint{0.000000in}{-0.048611in}}{\pgfqpoint{0.000000in}{0.000000in}}{%
\pgfpathmoveto{\pgfqpoint{0.000000in}{0.000000in}}%
\pgfpathlineto{\pgfqpoint{0.000000in}{-0.048611in}}%
\pgfusepath{stroke,fill}%
}%
\begin{pgfscope}%
\pgfsys@transformshift{0.877943in}{0.565000in}%
\pgfsys@useobject{currentmarker}{}%
\end{pgfscope}%
\end{pgfscope}%
\begin{pgfscope}%
\definecolor{textcolor}{rgb}{0.000000,0.000000,0.000000}%
\pgfsetstrokecolor{textcolor}%
\pgfsetfillcolor{textcolor}%
\pgftext[x=0.877943in,y=0.467777in,,top]{\color{textcolor}{\rmfamily\fontsize{10.000000}{12.000000}\selectfont\catcode`\^=\active\def^{\ifmmode\sp\else\^{}\fi}\catcode`\%=\active\def
\end{pgfscope}%
\begin{pgfscope}%
\pgfsetbuttcap%
\pgfsetroundjoin%
\definecolor{currentfill}{rgb}{0.000000,0.000000,0.000000}%
\pgfsetfillcolor{currentfill}%
\pgfsetlinewidth{0.803000pt}%
\definecolor{currentstroke}{rgb}{0.000000,0.000000,0.000000}%
\pgfsetstrokecolor{currentstroke}%
\pgfsetdash{}{0pt}%
\pgfsys@defobject{currentmarker}{\pgfqpoint{0.000000in}{-0.048611in}}{\pgfqpoint{0.000000in}{0.000000in}}{%
\pgfpathmoveto{\pgfqpoint{0.000000in}{0.000000in}}%
\pgfpathlineto{\pgfqpoint{0.000000in}{-0.048611in}}%
\pgfusepath{stroke,fill}%
}%
\begin{pgfscope}%
\pgfsys@transformshift{1.799009in}{0.565000in}%
\pgfsys@useobject{currentmarker}{}%
\end{pgfscope}%
\end{pgfscope}%
\begin{pgfscope}%
\definecolor{textcolor}{rgb}{0.000000,0.000000,0.000000}%
\pgfsetstrokecolor{textcolor}%
\pgfsetfillcolor{textcolor}%
\pgftext[x=1.799009in,y=0.467777in,,top]{\color{textcolor}{\rmfamily\fontsize{10.000000}{12.000000}\selectfont\catcode`\^=\active\def^{\ifmmode\sp\else\^{}\fi}\catcode`\%=\active\def
\end{pgfscope}%
\begin{pgfscope}%
\pgfsetbuttcap%
\pgfsetroundjoin%
\definecolor{currentfill}{rgb}{0.000000,0.000000,0.000000}%
\pgfsetfillcolor{currentfill}%
\pgfsetlinewidth{0.803000pt}%
\definecolor{currentstroke}{rgb}{0.000000,0.000000,0.000000}%
\pgfsetstrokecolor{currentstroke}%
\pgfsetdash{}{0pt}%
\pgfsys@defobject{currentmarker}{\pgfqpoint{0.000000in}{-0.048611in}}{\pgfqpoint{0.000000in}{0.000000in}}{%
\pgfpathmoveto{\pgfqpoint{0.000000in}{0.000000in}}%
\pgfpathlineto{\pgfqpoint{0.000000in}{-0.048611in}}%
\pgfusepath{stroke,fill}%
}%
\begin{pgfscope}%
\pgfsys@transformshift{2.720075in}{0.565000in}%
\pgfsys@useobject{currentmarker}{}%
\end{pgfscope}%
\end{pgfscope}%
\begin{pgfscope}%
\definecolor{textcolor}{rgb}{0.000000,0.000000,0.000000}%
\pgfsetstrokecolor{textcolor}%
\pgfsetfillcolor{textcolor}%
\pgftext[x=2.720075in,y=0.467777in,,top]{\color{textcolor}{\rmfamily\fontsize{10.000000}{12.000000}\selectfont\catcode`\^=\active\def^{\ifmmode\sp\else\^{}\fi}\catcode`\%=\active\def
\end{pgfscope}%
\begin{pgfscope}%
\definecolor{textcolor}{rgb}{0.000000,0.000000,0.000000}%
\pgfsetstrokecolor{textcolor}%
\pgfsetfillcolor{textcolor}%
\pgftext[x=1.799009in,y=0.288889in,,top]{\color{textcolor}{\rmfamily\fontsize{10.000000}{12.000000}\selectfont\catcode`\^=\active\def^{\ifmmode\sp\else\^{}\fi}\catcode`\%=\active\def
\end{pgfscope}%
\begin{pgfscope}%
\pgfsetbuttcap%
\pgfsetroundjoin%
\definecolor{currentfill}{rgb}{0.000000,0.000000,0.000000}%
\pgfsetfillcolor{currentfill}%
\pgfsetlinewidth{0.803000pt}%
\definecolor{currentstroke}{rgb}{0.000000,0.000000,0.000000}%
\pgfsetstrokecolor{currentstroke}%
\pgfsetdash{}{0pt}%
\pgfsys@defobject{currentmarker}{\pgfqpoint{-0.048611in}{0.000000in}}{\pgfqpoint{-0.000000in}{0.000000in}}{%
\pgfpathmoveto{\pgfqpoint{-0.000000in}{0.000000in}}%
\pgfpathlineto{\pgfqpoint{-0.048611in}{0.000000in}}%
\pgfusepath{stroke,fill}%
}%
\begin{pgfscope}%
\pgfsys@transformshift{0.785837in}{0.657106in}%
\pgfsys@useobject{currentmarker}{}%
\end{pgfscope}%
\end{pgfscope}%
\begin{pgfscope}%
\definecolor{textcolor}{rgb}{0.000000,0.000000,0.000000}%
\pgfsetstrokecolor{textcolor}%
\pgfsetfillcolor{textcolor}%
\pgftext[x=0.403120in, y=0.608912in, left, base]{\color{textcolor}{\rmfamily\fontsize{10.000000}{12.000000}\selectfont\catcode`\^=\active\def^{\ifmmode\sp\else\^{}\fi}\catcode`\%=\active\def
\end{pgfscope}%
\begin{pgfscope}%
\pgfsetbuttcap%
\pgfsetroundjoin%
\definecolor{currentfill}{rgb}{0.000000,0.000000,0.000000}%
\pgfsetfillcolor{currentfill}%
\pgfsetlinewidth{0.803000pt}%
\definecolor{currentstroke}{rgb}{0.000000,0.000000,0.000000}%
\pgfsetstrokecolor{currentstroke}%
\pgfsetdash{}{0pt}%
\pgfsys@defobject{currentmarker}{\pgfqpoint{-0.048611in}{0.000000in}}{\pgfqpoint{-0.000000in}{0.000000in}}{%
\pgfpathmoveto{\pgfqpoint{-0.000000in}{0.000000in}}%
\pgfpathlineto{\pgfqpoint{-0.048611in}{0.000000in}}%
\pgfusepath{stroke,fill}%
}%
\begin{pgfscope}%
\pgfsys@transformshift{0.785837in}{1.117639in}%
\pgfsys@useobject{currentmarker}{}%
\end{pgfscope}%
\end{pgfscope}%
\begin{pgfscope}%
\definecolor{textcolor}{rgb}{0.000000,0.000000,0.000000}%
\pgfsetstrokecolor{textcolor}%
\pgfsetfillcolor{textcolor}%
\pgftext[x=0.403120in, y=1.069445in, left, base]{\color{textcolor}{\rmfamily\fontsize{10.000000}{12.000000}\selectfont\catcode`\^=\active\def^{\ifmmode\sp\else\^{}\fi}\catcode`\%=\active\def
\end{pgfscope}%
\begin{pgfscope}%
\pgfsetbuttcap%
\pgfsetroundjoin%
\definecolor{currentfill}{rgb}{0.000000,0.000000,0.000000}%
\pgfsetfillcolor{currentfill}%
\pgfsetlinewidth{0.803000pt}%
\definecolor{currentstroke}{rgb}{0.000000,0.000000,0.000000}%
\pgfsetstrokecolor{currentstroke}%
\pgfsetdash{}{0pt}%
\pgfsys@defobject{currentmarker}{\pgfqpoint{-0.048611in}{0.000000in}}{\pgfqpoint{-0.000000in}{0.000000in}}{%
\pgfpathmoveto{\pgfqpoint{-0.000000in}{0.000000in}}%
\pgfpathlineto{\pgfqpoint{-0.048611in}{0.000000in}}%
\pgfusepath{stroke,fill}%
}%
\begin{pgfscope}%
\pgfsys@transformshift{0.785837in}{1.578172in}%
\pgfsys@useobject{currentmarker}{}%
\end{pgfscope}%
\end{pgfscope}%
\begin{pgfscope}%
\definecolor{textcolor}{rgb}{0.000000,0.000000,0.000000}%
\pgfsetstrokecolor{textcolor}%
\pgfsetfillcolor{textcolor}%
\pgftext[x=0.511145in, y=1.529978in, left, base]{\color{textcolor}{\rmfamily\fontsize{10.000000}{12.000000}\selectfont\catcode`\^=\active\def^{\ifmmode\sp\else\^{}\fi}\catcode`\%=\active\def
\end{pgfscope}%
\begin{pgfscope}%
\pgfsetbuttcap%
\pgfsetroundjoin%
\definecolor{currentfill}{rgb}{0.000000,0.000000,0.000000}%
\pgfsetfillcolor{currentfill}%
\pgfsetlinewidth{0.803000pt}%
\definecolor{currentstroke}{rgb}{0.000000,0.000000,0.000000}%
\pgfsetstrokecolor{currentstroke}%
\pgfsetdash{}{0pt}%
\pgfsys@defobject{currentmarker}{\pgfqpoint{-0.048611in}{0.000000in}}{\pgfqpoint{-0.000000in}{0.000000in}}{%
\pgfpathmoveto{\pgfqpoint{-0.000000in}{0.000000in}}%
\pgfpathlineto{\pgfqpoint{-0.048611in}{0.000000in}}%
\pgfusepath{stroke,fill}%
}%
\begin{pgfscope}%
\pgfsys@transformshift{0.785837in}{2.038705in}%
\pgfsys@useobject{currentmarker}{}%
\end{pgfscope}%
\end{pgfscope}%
\begin{pgfscope}%
\definecolor{textcolor}{rgb}{0.000000,0.000000,0.000000}%
\pgfsetstrokecolor{textcolor}%
\pgfsetfillcolor{textcolor}%
\pgftext[x=0.511145in, y=1.990511in, left, base]{\color{textcolor}{\rmfamily\fontsize{10.000000}{12.000000}\selectfont\catcode`\^=\active\def^{\ifmmode\sp\else\^{}\fi}\catcode`\%=\active\def
\end{pgfscope}%
\begin{pgfscope}%
\pgfsetbuttcap%
\pgfsetroundjoin%
\definecolor{currentfill}{rgb}{0.000000,0.000000,0.000000}%
\pgfsetfillcolor{currentfill}%
\pgfsetlinewidth{0.803000pt}%
\definecolor{currentstroke}{rgb}{0.000000,0.000000,0.000000}%
\pgfsetstrokecolor{currentstroke}%
\pgfsetdash{}{0pt}%
\pgfsys@defobject{currentmarker}{\pgfqpoint{-0.048611in}{0.000000in}}{\pgfqpoint{-0.000000in}{0.000000in}}{%
\pgfpathmoveto{\pgfqpoint{-0.000000in}{0.000000in}}%
\pgfpathlineto{\pgfqpoint{-0.048611in}{0.000000in}}%
\pgfusepath{stroke,fill}%
}%
\begin{pgfscope}%
\pgfsys@transformshift{0.785837in}{2.499238in}%
\pgfsys@useobject{currentmarker}{}%
\end{pgfscope}%
\end{pgfscope}%
\begin{pgfscope}%
\definecolor{textcolor}{rgb}{0.000000,0.000000,0.000000}%
\pgfsetstrokecolor{textcolor}%
\pgfsetfillcolor{textcolor}%
\pgftext[x=0.511145in, y=2.451044in, left, base]{\color{textcolor}{\rmfamily\fontsize{10.000000}{12.000000}\selectfont\catcode`\^=\active\def^{\ifmmode\sp\else\^{}\fi}\catcode`\%=\active\def
\end{pgfscope}%
\begin{pgfscope}%
\definecolor{textcolor}{rgb}{0.000000,0.000000,0.000000}%
\pgfsetstrokecolor{textcolor}%
\pgfsetfillcolor{textcolor}%
\pgftext[x=0.347564in,y=1.578172in,,bottom,rotate=90.000000]{\color{textcolor}{\rmfamily\fontsize{10.000000}{12.000000}\selectfont\catcode`\^=\active\def^{\ifmmode\sp\else\^{}\fi}\catcode`\%=\active\def
\end{pgfscope}%
\begin{pgfscope}%
\pgfsetrectcap%
\pgfsetmiterjoin%
\pgfsetlinewidth{0.803000pt}%
\definecolor{currentstroke}{rgb}{0.000000,0.000000,0.000000}%
\pgfsetstrokecolor{currentstroke}%
\pgfsetdash{}{0pt}%
\pgfpathmoveto{\pgfqpoint{0.785837in}{0.565000in}}%
\pgfpathlineto{\pgfqpoint{0.785837in}{2.591345in}}%
\pgfusepath{stroke}%
\end{pgfscope}%
\begin{pgfscope}%
\pgfsetrectcap%
\pgfsetmiterjoin%
\pgfsetlinewidth{0.803000pt}%
\definecolor{currentstroke}{rgb}{0.000000,0.000000,0.000000}%
\pgfsetstrokecolor{currentstroke}%
\pgfsetdash{}{0pt}%
\pgfpathmoveto{\pgfqpoint{2.812182in}{0.565000in}}%
\pgfpathlineto{\pgfqpoint{2.812182in}{2.591345in}}%
\pgfusepath{stroke}%
\end{pgfscope}%
\begin{pgfscope}%
\pgfsetrectcap%
\pgfsetmiterjoin%
\pgfsetlinewidth{0.803000pt}%
\definecolor{currentstroke}{rgb}{0.000000,0.000000,0.000000}%
\pgfsetstrokecolor{currentstroke}%
\pgfsetdash{}{0pt}%
\pgfpathmoveto{\pgfqpoint{0.785837in}{0.565000in}}%
\pgfpathlineto{\pgfqpoint{2.812182in}{0.565000in}}%
\pgfusepath{stroke}%
\end{pgfscope}%
\begin{pgfscope}%
\pgfsetrectcap%
\pgfsetmiterjoin%
\pgfsetlinewidth{0.803000pt}%
\definecolor{currentstroke}{rgb}{0.000000,0.000000,0.000000}%
\pgfsetstrokecolor{currentstroke}%
\pgfsetdash{}{0pt}%
\pgfpathmoveto{\pgfqpoint{0.785837in}{2.591345in}}%
\pgfpathlineto{\pgfqpoint{2.812182in}{2.591345in}}%
\pgfusepath{stroke}%
\end{pgfscope}%
\begin{pgfscope}%
\pgfsetbuttcap%
\pgfsetmiterjoin%
\definecolor{currentfill}{rgb}{1.000000,1.000000,1.000000}%
\pgfsetfillcolor{currentfill}%
\pgfsetlinewidth{0.000000pt}%
\definecolor{currentstroke}{rgb}{0.000000,0.000000,0.000000}%
\pgfsetstrokecolor{currentstroke}%
\pgfsetstrokeopacity{0.000000}%
\pgfsetdash{}{0pt}%
\pgfpathmoveto{\pgfqpoint{2.943367in}{0.565000in}}%
\pgfpathlineto{\pgfqpoint{3.044685in}{0.565000in}}%
\pgfpathlineto{\pgfqpoint{3.044685in}{2.591345in}}%
\pgfpathlineto{\pgfqpoint{2.943367in}{2.591345in}}%
\pgfpathlineto{\pgfqpoint{2.943367in}{0.565000in}}%
\pgfpathclose%
\pgfusepath{fill}%
\end{pgfscope}%
\begin{pgfscope}%
\pgfsys@transformshift{2.940000in}{0.569539in}%
\pgftext[left,bottom]{\includegraphics[interpolate=true,width=0.100000in,height=2.030000in]{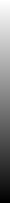}}%
\end{pgfscope}%
\begin{pgfscope}%
\pgfsetbuttcap%
\pgfsetroundjoin%
\definecolor{currentfill}{rgb}{0.000000,0.000000,0.000000}%
\pgfsetfillcolor{currentfill}%
\pgfsetlinewidth{0.803000pt}%
\definecolor{currentstroke}{rgb}{0.000000,0.000000,0.000000}%
\pgfsetstrokecolor{currentstroke}%
\pgfsetdash{}{0pt}%
\pgfsys@defobject{currentmarker}{\pgfqpoint{0.000000in}{0.000000in}}{\pgfqpoint{0.048611in}{0.000000in}}{%
\pgfpathmoveto{\pgfqpoint{0.000000in}{0.000000in}}%
\pgfpathlineto{\pgfqpoint{0.048611in}{0.000000in}}%
\pgfusepath{stroke,fill}%
}%
\begin{pgfscope}%
\pgfsys@transformshift{3.044685in}{0.565000in}%
\pgfsys@useobject{currentmarker}{}%
\end{pgfscope}%
\end{pgfscope}%
\begin{pgfscope}%
\definecolor{textcolor}{rgb}{0.000000,0.000000,0.000000}%
\pgfsetstrokecolor{textcolor}%
\pgfsetfillcolor{textcolor}%
\pgftext[x=3.141907in, y=0.516805in, left, base]{\color{textcolor}{\rmfamily\fontsize{10.000000}{12.000000}\selectfont\catcode`\^=\active\def^{\ifmmode\sp\else\^{}\fi}\catcode`\%=\active\def
\end{pgfscope}%
\begin{pgfscope}%
\pgfsetbuttcap%
\pgfsetroundjoin%
\definecolor{currentfill}{rgb}{0.000000,0.000000,0.000000}%
\pgfsetfillcolor{currentfill}%
\pgfsetlinewidth{0.803000pt}%
\definecolor{currentstroke}{rgb}{0.000000,0.000000,0.000000}%
\pgfsetstrokecolor{currentstroke}%
\pgfsetdash{}{0pt}%
\pgfsys@defobject{currentmarker}{\pgfqpoint{0.000000in}{0.000000in}}{\pgfqpoint{0.048611in}{0.000000in}}{%
\pgfpathmoveto{\pgfqpoint{0.000000in}{0.000000in}}%
\pgfpathlineto{\pgfqpoint{0.048611in}{0.000000in}}%
\pgfusepath{stroke,fill}%
}%
\begin{pgfscope}%
\pgfsys@transformshift{3.044685in}{0.970269in}%
\pgfsys@useobject{currentmarker}{}%
\end{pgfscope}%
\end{pgfscope}%
\begin{pgfscope}%
\definecolor{textcolor}{rgb}{0.000000,0.000000,0.000000}%
\pgfsetstrokecolor{textcolor}%
\pgfsetfillcolor{textcolor}%
\pgftext[x=3.141907in, y=0.922074in, left, base]{\color{textcolor}{\rmfamily\fontsize{10.000000}{12.000000}\selectfont\catcode`\^=\active\def^{\ifmmode\sp\else\^{}\fi}\catcode`\%=\active\def
\end{pgfscope}%
\begin{pgfscope}%
\pgfsetbuttcap%
\pgfsetroundjoin%
\definecolor{currentfill}{rgb}{0.000000,0.000000,0.000000}%
\pgfsetfillcolor{currentfill}%
\pgfsetlinewidth{0.803000pt}%
\definecolor{currentstroke}{rgb}{0.000000,0.000000,0.000000}%
\pgfsetstrokecolor{currentstroke}%
\pgfsetdash{}{0pt}%
\pgfsys@defobject{currentmarker}{\pgfqpoint{0.000000in}{0.000000in}}{\pgfqpoint{0.048611in}{0.000000in}}{%
\pgfpathmoveto{\pgfqpoint{0.000000in}{0.000000in}}%
\pgfpathlineto{\pgfqpoint{0.048611in}{0.000000in}}%
\pgfusepath{stroke,fill}%
}%
\begin{pgfscope}%
\pgfsys@transformshift{3.044685in}{1.375538in}%
\pgfsys@useobject{currentmarker}{}%
\end{pgfscope}%
\end{pgfscope}%
\begin{pgfscope}%
\definecolor{textcolor}{rgb}{0.000000,0.000000,0.000000}%
\pgfsetstrokecolor{textcolor}%
\pgfsetfillcolor{textcolor}%
\pgftext[x=3.141907in, y=1.327343in, left, base]{\color{textcolor}{\rmfamily\fontsize{10.000000}{12.000000}\selectfont\catcode`\^=\active\def^{\ifmmode\sp\else\^{}\fi}\catcode`\%=\active\def
\end{pgfscope}%
\begin{pgfscope}%
\pgfsetbuttcap%
\pgfsetroundjoin%
\definecolor{currentfill}{rgb}{0.000000,0.000000,0.000000}%
\pgfsetfillcolor{currentfill}%
\pgfsetlinewidth{0.803000pt}%
\definecolor{currentstroke}{rgb}{0.000000,0.000000,0.000000}%
\pgfsetstrokecolor{currentstroke}%
\pgfsetdash{}{0pt}%
\pgfsys@defobject{currentmarker}{\pgfqpoint{0.000000in}{0.000000in}}{\pgfqpoint{0.048611in}{0.000000in}}{%
\pgfpathmoveto{\pgfqpoint{0.000000in}{0.000000in}}%
\pgfpathlineto{\pgfqpoint{0.048611in}{0.000000in}}%
\pgfusepath{stroke,fill}%
}%
\begin{pgfscope}%
\pgfsys@transformshift{3.044685in}{1.780807in}%
\pgfsys@useobject{currentmarker}{}%
\end{pgfscope}%
\end{pgfscope}%
\begin{pgfscope}%
\definecolor{textcolor}{rgb}{0.000000,0.000000,0.000000}%
\pgfsetstrokecolor{textcolor}%
\pgfsetfillcolor{textcolor}%
\pgftext[x=3.141907in, y=1.732612in, left, base]{\color{textcolor}{\rmfamily\fontsize{10.000000}{12.000000}\selectfont\catcode`\^=\active\def^{\ifmmode\sp\else\^{}\fi}\catcode`\%=\active\def
\end{pgfscope}%
\begin{pgfscope}%
\pgfsetbuttcap%
\pgfsetroundjoin%
\definecolor{currentfill}{rgb}{0.000000,0.000000,0.000000}%
\pgfsetfillcolor{currentfill}%
\pgfsetlinewidth{0.803000pt}%
\definecolor{currentstroke}{rgb}{0.000000,0.000000,0.000000}%
\pgfsetstrokecolor{currentstroke}%
\pgfsetdash{}{0pt}%
\pgfsys@defobject{currentmarker}{\pgfqpoint{0.000000in}{0.000000in}}{\pgfqpoint{0.048611in}{0.000000in}}{%
\pgfpathmoveto{\pgfqpoint{0.000000in}{0.000000in}}%
\pgfpathlineto{\pgfqpoint{0.048611in}{0.000000in}}%
\pgfusepath{stroke,fill}%
}%
\begin{pgfscope}%
\pgfsys@transformshift{3.044685in}{2.186076in}%
\pgfsys@useobject{currentmarker}{}%
\end{pgfscope}%
\end{pgfscope}%
\begin{pgfscope}%
\definecolor{textcolor}{rgb}{0.000000,0.000000,0.000000}%
\pgfsetstrokecolor{textcolor}%
\pgfsetfillcolor{textcolor}%
\pgftext[x=3.141907in, y=2.137881in, left, base]{\color{textcolor}{\rmfamily\fontsize{10.000000}{12.000000}\selectfont\catcode`\^=\active\def^{\ifmmode\sp\else\^{}\fi}\catcode`\%=\active\def
\end{pgfscope}%
\begin{pgfscope}%
\pgfsetbuttcap%
\pgfsetroundjoin%
\definecolor{currentfill}{rgb}{0.000000,0.000000,0.000000}%
\pgfsetfillcolor{currentfill}%
\pgfsetlinewidth{0.803000pt}%
\definecolor{currentstroke}{rgb}{0.000000,0.000000,0.000000}%
\pgfsetstrokecolor{currentstroke}%
\pgfsetdash{}{0pt}%
\pgfsys@defobject{currentmarker}{\pgfqpoint{0.000000in}{0.000000in}}{\pgfqpoint{0.048611in}{0.000000in}}{%
\pgfpathmoveto{\pgfqpoint{0.000000in}{0.000000in}}%
\pgfpathlineto{\pgfqpoint{0.048611in}{0.000000in}}%
\pgfusepath{stroke,fill}%
}%
\begin{pgfscope}%
\pgfsys@transformshift{3.044685in}{2.591345in}%
\pgfsys@useobject{currentmarker}{}%
\end{pgfscope}%
\end{pgfscope}%
\begin{pgfscope}%
\definecolor{textcolor}{rgb}{0.000000,0.000000,0.000000}%
\pgfsetstrokecolor{textcolor}%
\pgfsetfillcolor{textcolor}%
\pgftext[x=3.141907in, y=2.543150in, left, base]{\color{textcolor}{\rmfamily\fontsize{10.000000}{12.000000}\selectfont\catcode`\^=\active\def^{\ifmmode\sp\else\^{}\fi}\catcode`\%=\active\def
\end{pgfscope}%
\begin{pgfscope}%
\pgfsetrectcap%
\pgfsetmiterjoin%
\pgfsetlinewidth{0.803000pt}%
\definecolor{currentstroke}{rgb}{0.000000,0.000000,0.000000}%
\pgfsetstrokecolor{currentstroke}%
\pgfsetdash{}{0pt}%
\pgfpathmoveto{\pgfqpoint{2.943367in}{0.565000in}}%
\pgfpathlineto{\pgfqpoint{2.994026in}{0.565000in}}%
\pgfpathlineto{\pgfqpoint{3.044685in}{0.565000in}}%
\pgfpathlineto{\pgfqpoint{3.044685in}{2.591345in}}%
\pgfpathlineto{\pgfqpoint{2.994026in}{2.591345in}}%
\pgfpathlineto{\pgfqpoint{2.943367in}{2.591345in}}%
\pgfpathlineto{\pgfqpoint{2.943367in}{0.565000in}}%
\pgfpathclose%
\pgfusepath{stroke}%
\end{pgfscope}%
\end{pgfpicture}%
\makeatother%
\endgroup%